\documentclass[10pt]{IEEEtran}
\usepackage{multirow}
\usepackage{amsmath, amssymb, graphicx, booktabs, hyperref, geometry}
\usepackage{array}
\usepackage{tabularx}
\usepackage{float}
\usepackage{longtable} 
\usepackage{etoolbox}
\usepackage{pdflscape}
\usepackage{caption}
\usepackage{svg}
\usepackage{setspace}
\usepackage{stfloats}
\usepackage{orcidlink}
\usepackage{comment}
\usepackage{amssymb}
\usepackage{enumitem}

\usepackage[font=scriptsize,skip=0pt]{caption}
\usepackage{xcolor}
\def\BibTeX{{\rm B\kern-.05em{\sc i\kern-.025em b}\kern-.08em
    T\kern-.1667em\lower.7ex\hbox{E}\kern-.125emX}}
\usepackage{cite}
\usepackage{braket}
\usepackage{adjustbox}
\usepackage{siunitx} 
 \renewcommand{\arraystretch}{1.2}
\begin{document}
\title{Quantum vs. Classical Machine Learning: \\ A Benchmark Study for Financial Prediction}
\author{\IEEEauthorblockN{Rehan Ahmad\textsuperscript{1}, Muhammad Kashif\textsuperscript{2,3}, Nouhaila Innan\textsuperscript{2,3}, and Muhammad Shafique\textsuperscript{2,3}\\}
\IEEEauthorblockA{
\textsuperscript{1}Department of Physics, Lahore University of Management Sciences (LUMS), Lahore Pakistan\\
\textsuperscript{2}eBRAIN Lab, Division of Engineering, New York University Abu Dhabi (NYUAD), Abu Dhabi, UAE\\
\textsuperscript{3}Center for Quantum and Topological Systems (CQTS), NYUAD Research Institute, NYUAD, Abu Dhabi, UAE\\
Emails: \{24100219@lums.edu.pk\},\{muhammadkashif, nouhaila.innan, muhammad.shafique\}@nyu.edu\\
\vspace{-20pt}
}}
\maketitle

\begin{abstract}


In this paper, we present a reproducible benchmarking framework that systematically compares QML models with architecture-matched classical counterparts across three financial tasks: (i) directional return prediction on U.S. and Turkish equities, (ii) live-trading simulation with Quantum LSTMs versus classical LSTMs on the S\&P 500, and (iii) realized volatility forecasting using Quantum Support Vector Regression. By standardizing data splits, features, and evaluation metrics, our study provides a fair assessment of when current-generation QML models can match or exceed classical methods. 

Our results reveal that quantum approaches show performance gains when data structure and circuit design are well aligned. In directional classification, hybrid quantum neural networks surpass the parameter-matched ANN by \textbf{+3.8 AUC} and \textbf{+3.4 accuracy points} on \texttt{AAPL} stock and by \textbf{+4.9 AUC} and \textbf{+3.6 accuracy points} on Turkish stock \texttt{KCHOL}. In live trading, the QLSTM achieves higher risk-adjusted returns in \textbf{two of four} S\&P~500 regimes. For volatility forecasting, an angle-encoded QSVR attains the \textbf{lowest QLIKE} on \texttt{KCHOL} and remains within $\sim$0.02--0.04 QLIKE of the best classical kernels on \texttt{S\&P~500} and \texttt{AAPL}. Our benchmarking framework clearly identifies the scenarios where current QML architectures offer tangible improvements and where established classical methods continue to dominate.

\end{abstract}

\section{Introduction}

The challenges of forecasting financial markets are rooted in market efficiency theories and statistical properties of financial data. The Efficient Market Hypothesis (EMH) asserts that asset prices fully reflect all available information \cite{fama1970efficient}. In its weak form, the hypothesis states that past price and volume data have no predictive power over future prices. In its strong form, it argues that even insider information is already incorporated into market prices. 
Despite the strong claims of EMH, a substantial body of research documents patterns that deviate from its predictions. The seminal work by Lo \& MacKinlay (1988) \cite{lo1988stock} uncovers short-term autocorrelations in stock returns that deviate from random walks. Similarly, Shiller (2003) \cite{shiller2003efficient} argues that behavioral biases can lead to effects like overconfidence and herding, which can cause deviation from EMH. Besides, empirical research supports that developing markets often diverge from the EMH hypothesis due to factors such as lower liquidity and higher transaction costs \cite{balaban1996stock}. 

These observations motivate classical statistical approaches that aim to model the probabilistic structure of financial time series. Factor models ranging from the classic Capital Asset Pricing Model (CAPM) \cite{sharpe1964capital} to multi‐factor extensions such as the Fama–French three‐factor model \cite{fama1993common} seek to explain cross‐sectional return through exposures to systematic risk factors. For volatility modeling, the Autoregressive Conditional Heteroskedasticity (ARCH) framework \cite{engle1982arch} and its generalization, the Generalized ARCH (GARCH) \cite{bollerslev1986garch} allow the conditional variance of return to evolve over time.  These methods provide a transparent statistical foundation for financial forecasting and continue to yield important insights into market behavior. However, they typically rely on linear relationships and may struggle to capture nonlinear interactions observed in real‐world markets.

\begin{figure*}
    \centering
    \includegraphics[width=1.0\linewidth]{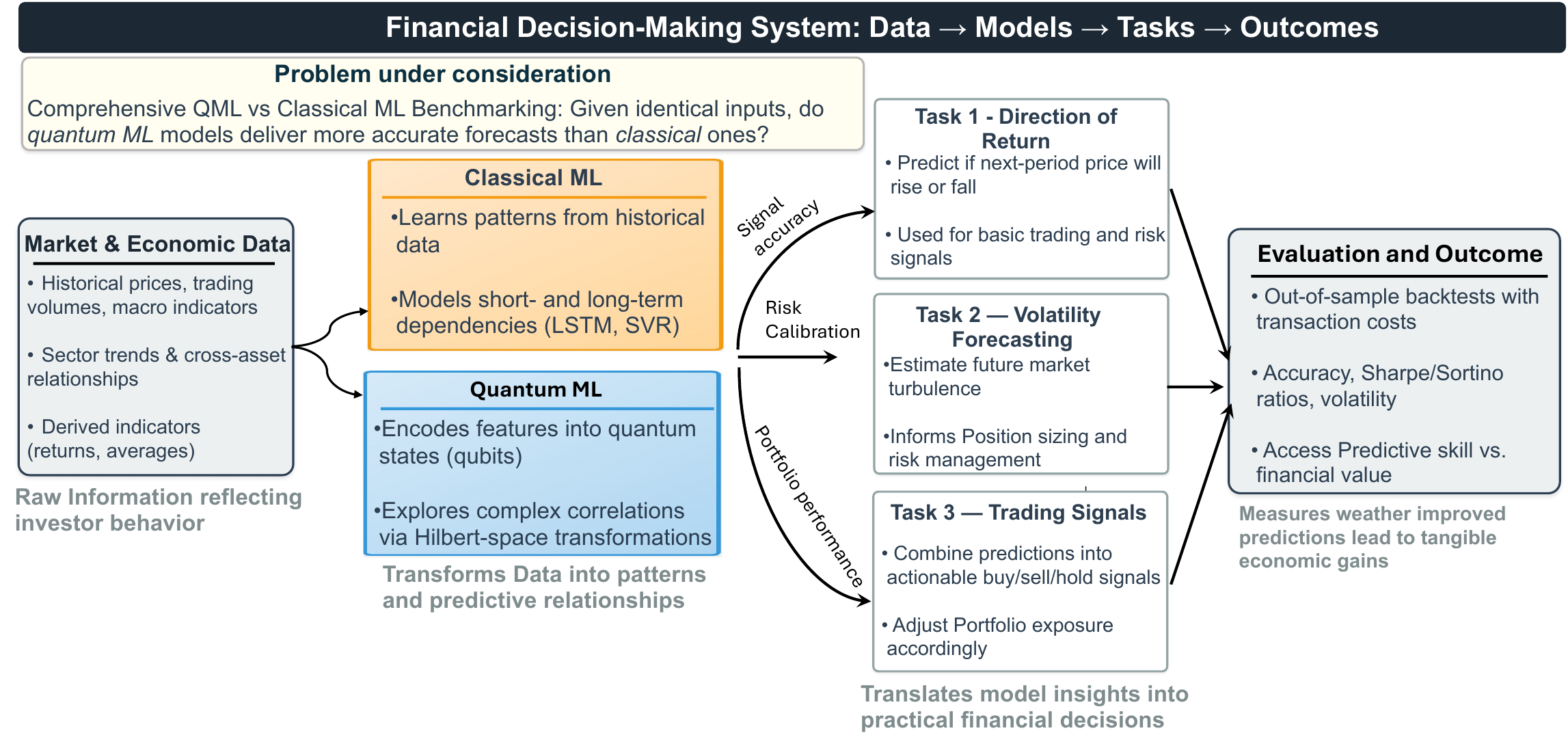}
    \caption{Overview of financial decision system, motivations and contributions of our work.}
    \label{fig:contributions}
\end{figure*}

To address these limitations, Machine Learning (ML) techniques have been increasingly adopted in finance due to their nonparametric nature. Prior works using Support Vector Machines (SVMs) demonstrated that nonlinear kernels can exploit technical indicators and lagged features to produce returns forecasts that outperform random‐walk benchmarks \cite{henrique2018stock}. Random forests and gradient boosting machines further improve predictive accuracy by aggregating simple decision trees. These ensemble methods flexibly capture higher‐order interactions between features \cite{ballings2015evaluating}. More recently, deep learning methods such as Long Short‐Term Memory (LSTM) networks have been applied to model nonlinear dependencies and long‐range temporal structures in financial markets. For example, the seminal work of Fischer \& Krauss (2018) \cite{fischer2018deep} show that LSTMs trained on rolling windows of S\&P 500 constituent returns achieve statistically significant gains in cross-sectional returns relative to logistic regression. Despite these advances, ML models are not without their tradeoffs: they can suffer from overfitting in low‐data regimes, struggle to generalize under drastic market shifts, and often require extensive feature engineering.

These challenges have motivated the exploration of alternative computational paradigms. In recent years, Quantum Machine Learning (QML) has emerged as a novel framework that may offer unique advantages to financial applications \cite{pathak2024resource,innan2024financial,innan2025qfnn,innan2024financial1,kashif2025evaluating,innan2025quantum,sawaika2025privacy,innan2025circuithunt,alami2024comparative,innan2024lep,alami2025fid,khan2025iqnn}, particularly in financial forecasting \cite{herman2022survey,orus2019quantum,kobayashi2023cross,kea2024hybrid,dutta2024qadqn,choudhary2025hqnn}. By encoding classical data into quantum states and leveraging the high‐dimensional Hilbert space of quantum systems \cite{kashif:2022_demonstrating,kashif:2025_computational}, QML models such as Quantum Neural Networks (QNNs), Quantum Long Short-Term Memory (QLSTM), and Quantum Support Vector Regression (QSVR) seek to capture complex patterns with fewer parameters than their classical counterparts \cite{kashif2021design,rebentrost2014quantum,suzuki2024quantum,innan2023enhancing,zaman2024studying,innan2024variational,innan2025next}. Prior studies have explored QNNs for return regression tasks \cite{emmanoulopoulos2022quantum}, but often fall short by omitting rich feature sets, such as technical indicators, or by comparing against mismatched baselines. Moreover, quantum recurrent architectures trained on only self-lag returns collapse into trivial persistence models unless they incorporate sufficient exogenous features \cite{li2023quantum}. Similarly, while Support Vector Regression (SVR) has long been applied to volatility forecasting, Quantum Support Vector Regression (QSVR) remains largely unexplored for this task, with most studies evaluating it on generic regression benchmarks rather than realized-volatility settings. Consequently, it remains unclear in which regimes QML can deliver improvements over well-tuned classical methods for key financial tasks such as return directional classification, live-trading, and volatility forecasting.


\subsection{Our Contributions}
To address the limitations of existing financial modeling approaches, our study develops a reproducible benchmarking framework to systematically compare QML models with architecture-matched classical counterparts that share similar inductive biases. We focus on three core financial tasks across equity markets of different maturity, using data from the developed U.S. market and the emerging Turkish market. By standardizing data splits, feature sets, and evaluation metrics, we aim to provide fair assessment of when, and under what conditions, current-generation QML models can outperform or match their classical baselines. An overview of financial decision system along with the motivation driving this study, and scope of our contributions, is summarized in Fig.~\ref{fig:contributions}, and explained as follows:

\begin{itemize}[left=0pt]
    \item \textbf{Directional Return Prediction.} We curate a cross-market dataset comprising Turkish and U.S. equities and evaluate QNNs against Artificial Neural Networks (ANNs) to predict next-day return direction. We progressively scale feature complexity from low-dimensional technical indicators (scarce and noisy data) to macroeconomic features for the S\&P 500, and finally to high-dimensional inputs for large-cap U.S. equities. This design probes how QNNs compare to ANNs across regimes of feature space growth and market maturity.
    
    \item \textbf{Live-Trading Simulation.} We benchmark QLSTM networks against classical LSTMs on the S\&P 500 index using identical rolling windows and walk-forward splits. By translating model probabilities into long–short trading strategies, we evaluate out-of-sample performance of both approaches using risk-adjusted performance and drawdown-aware metrics, thereby assessing the practical value of QML models in strategy execution.
    
    \item \textbf{Volatility Forecasting.} We compare QSVR with tuned classical SVR models for one-day-ahead realized volatility prediction. By holding the feature set constant, we isolate whether the quantum kernel embedding in Hilbert space yields better forecasting accuracy and risk-management utility in moderate-to high-dimensional settings.
    
    \item \textbf{Benchmark Design.} Beyond the individual case studies, we contribute a standardized benchmark framework for evaluating QML in finance. This includes aligned data splits, matched model architectures, and consistent evaluation metrics, providing the community with a reproducible foundation for future research in financial forecasting.
\end{itemize}

The remainder of this paper is organized as follows. 
Section~\ref{sec:Background} reviews the theoretical background of financial forecasting and surveys related work in classical and quantum ML. 
Section~\ref{sec:Methodology} outlines our unified benchmarking framework, including datasets, model architectures, and evaluation protocols. 
Section~\ref{sec:ResultsDiscussion} presents our experimental results and discussion across three tasks: (i) directional classification with QNNs versus ANNs, (ii) live trading simulations with QLSTMs versus LSTMs, and (iii) volatility forecasting with QSVRs against SVR and GARCH. 
Finally, Section~\ref{sec:Conclusion} summarizes our findings and outlines directions for future research.

\section{Background and Related Work \label{sec:Background}}
\subsection{Preliminaries}

\subsubsection{Directional Classification}
In the domain of price‐versus‐returns modeling, directional classification is particularly appealing. Rather than predicting the magnitude of today’s return, directional models cast forecasting as a binary decision: will today’s closing price exceed yesterday’s? The binary target variable $y_{t}$ is specified as:

\begin{equation}
    y_{t} \;=\;
    \begin{cases}
        1, & P_{t} > P_{t-1},\\
        0, & P_{t} \le P_{t-1}.
    \end{cases}
    \label{eq:binary_target}
\end{equation}

where $P_{t}$ is the asset price at time $t$, and $P_{t-1}$ is the asset price at the previous time step.

Directional classification emphasizes the \emph{sign} of returns rather than their magnitude. Learning an up/down signal is often easier than predicting precise return levels, since even small regression errors around zero can flip the sign and cause losing trades. By contrast, classification directly optimizes directional accuracy by aligning the loss function more closely with trading actions (long versus short). Empirical evidence supports this design choice: Leung, Daouk, and Chen (2000) \cite{leung2000forecasting} show that predicting the direction of index returns often produces more consistent trading performance than forecasting return values. As a result, directional models typically deliver more stable real‐world outcomes.

\subsubsection{Live Trading Simulation}

While predictive accuracy is informative, it does not directly translate into profitable financial outcomes. Live trading simulation or \textit{backtesting} provides the crucial bridge by transforming forecasts into trading decisions. In this framework, model predictions determine whether a position is long, short, or neutral, and realized returns are accumulated over time. 
Let $\hat{y}_{t}$ denote the trading position implied by the forecast at time $t$ and $r_{t+1}$ the realized return. The simulated strategy return is:
\begin{equation}
R_{t+1}^{\text{strategy}} = \hat{y}_{t} \cdot r_{t+1},
\label{eq:trading_return_simple}
\end{equation}
which may be adjusted to account for transaction costs. Portfolio wealth then evolves recursively as $W_{t} = W_{t-1}(1 + R_{t}^{\text{strategy}})$, producing an \emph{equity curve} that can be compared to a passive buy-and-hold baseline.
Through this lens, live trading simulation evaluates models not only on statistical fit but also on economic utility, offering a practical measure of whether predictive skill yields sustainable gains under market frictions.

\subsubsection{Volatility Forecasting}
\begin{equation}
\sigma_{\text{daily}} = \sqrt{ \frac{1}{N - 1} \sum_{t = 1}^{N} \left( r_t - \bar{r} \right)^2 }, 
\quad \text{where} \quad 
\bar{r} = \frac{1}{N} \sum_{t = 1}^{N} r_t.
\label{eq:daily_volatility}
\end{equation}

In financial markets, Mandelbrot (1963) \cite{Mandelbrot1963} was the first to observe that large changes tend to be followed by large changes and small changes by small changes regardless of the direction. This phenomenon, known as volatility clustering \cite{cont2007volatility}, means that standard linear time‐series models like Autoregressive Integrated Moving Average (ARIMA) \cite{box1976analysis}  fail to capture the changing variance over time since they typically assume homoscedastic errors. Engle (1982) \cite{engle1982arch} was the first one to accommodate for the conditional variance of volatility by introducing ARCH. Bollerslev (1986) \cite{bollerslev1986garch} later generalized ARCH to the GARCH. Over time, researchers have addressed numerous extensions to address asymmetries and tail behavior of these models, such as Exponential GARCH \cite{nelson1991egarch} and GJR‐GARCH \cite{glosten1993gjr}.

\subsection{Related Work}
\subsubsection{Machine Learning in Finance}

Machine learning (ML) methods, unlike traditional time-series approaches, are data-driven and non-parametric, allowing them to capture nonlinear financial relationships without explicit functional forms. Studies such as \cite{henrique2018stock} and \cite{ballings2015evaluating} show that SVMs, Random Forests, and k-NNs outperform linear models, though their success depends heavily on feature design and market stability.
Deep learning, particularly LSTMs, has shown strong performance in financial forecasting \cite{zou2022stock}. LSTMs trained on large datasets, such as Chinese market returns \cite{chen2015lstm} or S\&P 500 constituents \cite{fischer2018deep}, achieve statistically significant improvements in directional accuracy, while later works \cite{moghar2020stock,nguyen2022efficiency} extend these results to trading strategies through backtesting.


Moreover, properly tuned classical ML models typically achieve 55–65\% directional accuracy \cite{guyard2024predicting,zhong2017forecasting}, with cross-asset features further improving the performance \cite{jiao2017predicting}. In emerging markets where the EMH is weaker, such as the Tehran Stock Exchange, studies report directional classification accuracies that surpass those observed in developed markets. This suggests that greater informational inefficiencies can provide stronger predictive signals \cite{senol2009stock}, \cite{nabipour2020predicting}.
%
For volatility forecasting, SVM-based models serve as nonparametric alternatives to GARCH, often improving out-of-sample performance \cite{chen2010forecasting,perez2003estimating}, underscoring the data-driven advantage of ML in risk estimation.


\subsubsection{Quantum Machine Learning in Finance} \label{sec: QML for finance}
Quantum algorithms and QML are increasingly being explored in finance, motivated by the potential to capture high-dimensional dependencies and nonlinearities beyond classical methods\cite{zaman:2024_POQA,kashif2025evaluating}. However, prior works \cite{emmanoulopoulos2022quantum} often fall short in constructing expressive features spaces that can be utilized by quantum models to extract signals on downstream tasks. The primary reason behind this is that they omit valuable features, such as technical indicators and lagged values of other equities, which can serve as informative signals in financial forecasting. Furthermore, these studies frequently compare quantum models against structurally inappropriate classical baselines. In \cite{emmanoulopoulos2022quantum}, a shallow QNN is compared with an LSTM, which is an architecture \textit{specifically} designed to model long-range temporal dependencies.  
Similarly, existing studies in QML for finance often restrict forecasting to point-value regression, assessing performance only through standard error metrics such as RMSE, MAE, or $R^2$ \cite{kea2024hybrid}. These scalars do not translate cleanly into actionable trade rules. Our approach utilizes quantum models to produce a discrete directional output that traders can directly use to generate buy/sell decisions. This makes the model performance more interpretable. 

Furthermore, when Quantum RNNs or QLSTMs are trained on very short lags such as the seven-day window as in \cite{li2023quantum}, they risk degenerating into \textit{persistence models}: outputting yesterday’s price yields deceptively low mean-square error yet providing a catastrophic trading signal. Incorporation of a large number of exogenous features or trading sequences going beyond a single stock is required to overcome this limitation. However, here, the capacity of current noisy quantum hardware becomes a constraint. For example, the classical LSTM benchmark of \cite{fischer2018deep} processes hundreds of thousands of overlapping 200-day return sequences. A straightforward QLSTM \cite{chen2022quantum} implementation of the same experiment would require millions of circuit executions which is far beyond the capacity of current noisy quantum hardware.
Similar scalability bottlenecks appear for high-dimensional inputs. The authors in \cite{jiao2017predicting} use a diverse set of inputs, such as entropy, energy, and kurtosis, to obtain tradable signals using ANNs. However, their feature space, which totals a size of 223, is well beyond the capacity of current QNNs if angle-embedding is employed. Given this, a central part of our study is how we can adapt QNNs to perform well on different dimensional regimes. A noteworthy work in this regard is \cite{zhong2017forecasting}, which utilizes data compression techniques such as PCA before sending it through a simple singled layered ANN. For QNNs, we analyze how to cope with different dimensionalities and whether different embedding techniques, such as amplitude encoding, can be employed without compromising on overall accuracy. 

Finally, despite its theoretical formulation \cite{rebentrost2014quantum,suzuki2024quantum}, Quantum Support Vector Regression (QSVR) remains largely \emph{unexplored} for realized-volatility forecasting. Prior studies have primarily examined generic regression or classification tasks, leaving unanswered whether quantum kernels can meaningfully improve volatility-aware metrics such as QLIKE or achieve statistically significant gains under Diebold–Mariano (DM) tests. Consequently, it remains unclear under what data regimes quantum kernels can outperform optimized classical counterparts

Overall, our benchmark study clarifies where QML can deliver improvements, if any, over established ML methods in the context of return prediction and risk forecasting. We summarize key gaps in prior work and our remedies in Table~\ref{tab:sota_summary}.

\begin{table*}[t]
  \centering
  \footnotesize
  \caption{\scriptsize Representative prior QML-in-finance studies and how our benchmark addresses their limitations.}
  \label{tab:sota_summary}
  \renewcommand{\arraystretch}{1.25}
  \begin{adjustbox}{max width=\textwidth}
  \begin{tabularx}{\textwidth}{@{} >{\raggedright\arraybackslash}X
                                    >{\raggedright\arraybackslash}l
                                    >{\raggedright\arraybackslash}X
                                    >{\raggedright\arraybackslash}X
                                    >{\raggedright\arraybackslash}X @{}}
    \toprule
    \textbf{Study} & \textbf{Method} & \textbf{Task / Dataset} & \textbf{Key Limitation} & \textbf{How Our Benchmark Improves} \\
    \midrule
    Emmanoulopoulos et al.\ (2022) \cite{emmanoulopoulos2022quantum}
    & QNN
    & Return regression (stocks)
    & Lacks rich features; mismatched baseline vs.\ LSTM
    & Uses feature-rich datasets and architecture-matched QNN–ANN comparison \\
    
    Li et al.\ (2023) \cite{li2023quantum}
    & QRNN
    & Sequence forecasting using past returns as sole feature
    & Degenerates into persistence model
    & Incorporates multi-feature windows for realistic trading evaluation \\
    
    Kea et al.\ (2024) \cite{kea2024hybrid}
    & Hybrid QLSTM
    & Regression (synthetic)
    & Evaluates only statistical fit (RMSE)
    & Extends evaluation to directional accuracy, risk-adjusted return, and QLIKE for finance \\
    
    Li, Mukhopadhyay, Bayat \& Habibnia (2025) \cite{li2025quantum}
    & QRC
    & Realized volatility forecasting (\texttt{S\&P 500}, monthly)
    & Evaluates only on a single asset, lacks matched baselines and formal statistical tests (QLIKE, DM)
    & Multi-asset (U.S./Turkey), architecture-matched QSVR vs.\ SVR/GARCH with QLIKE and DM tests \\
    
    \textbf{Our Work (2025)}
    & \textbf{Unified QML Benchmark}
    & \textbf{Directional Classification, Live Trading, and Volatility Forecasting (U.S./Turkey)}
    & \textbf{Previous works task-specific or non-standardized}
    & \textbf{Unifies tasks, features, architectures, and metrics; introduces large-scale QSVR volatility benchmarking} \\
    \bottomrule
  \end{tabularx}
  \end{adjustbox}
\end{table*}

\begin{figure*}
    \centering
    \includegraphics[width=1.0\textwidth]{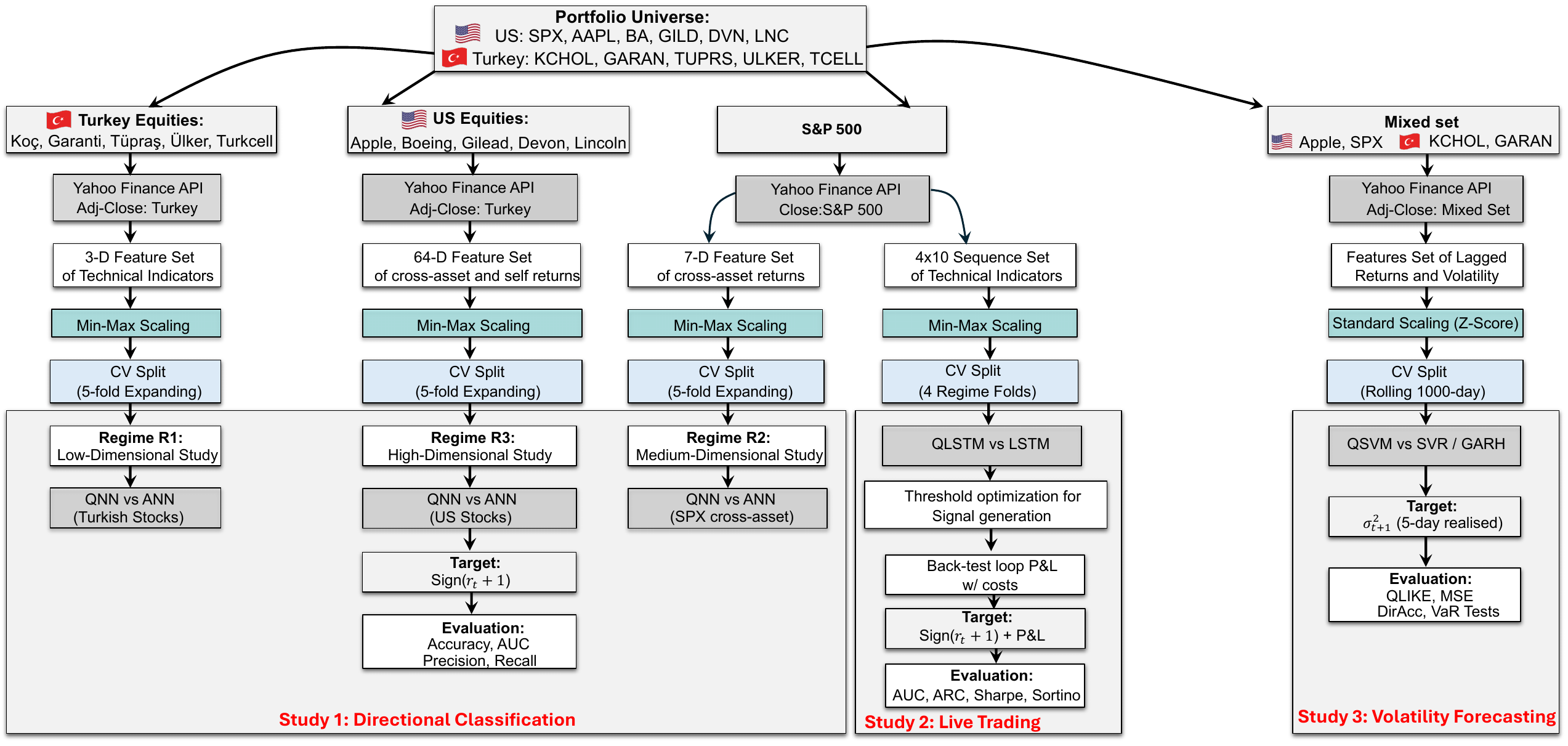}
    \caption{\scriptsize  \textbf{A detailed overview of our methodology}. We assemble a \emph{Portfolio Universe} that spans U.S and Turkey Markets and then fetch daily prices with the Yahoo Finance API. Each branch applies task‑specific \textit{feature engineering}. Features are subsequently scaled and split by cross‑validation scheme before being fed into paired quantum and classical models: QNN vs ANN, QLSTM vs LSTM, and QSVR vs SVR. Downstream arrows indicate the learning target for each study and the evaluation metrics used to compare models.}
    \label{fig:revised_pipeline}
\end{figure*}

\section{Our Methodology} \label{sec:Methodology}

In this paper, we perform a comprehensive benchmarking analysis of different QML models on a variety of important stock market forecasting problems and compare their performance against classical ML counterparts. A detailed overview of our methodology is presented in Fig. \ref{fig:revised_pipeline}, and a step-by-step ooverview is below:

\begin{itemize}
    \item \textbf{Objective:} Conduct a comprehensive benchmarking of diverse QML models on key stock market forecasting tasks.

    \item \textbf{Comparison Baselines:} Evaluate QML models against classical ML counterparts, including neural- and kernel-based architectures.

    \item \textbf{Standardization:} Use consistent data splits across all experiments, apply unified hyperparameter tuning protocols, and employ common evaluation metrics for fair comparison.

    \item \textbf{Optimization Analysis:} Identify optimal QML configurations in terms of qubit count and circuit depth, and determine performance sweet spots where QML achieves best performance.

    \item \textbf{Outcome:} Provide actionable insights for QML researchers and financial practitioners, and highlight scenarios where quantum models outperform classical baselines.
\end{itemize}


\subsection{Task-Definition and Experimental Scope }

Our methodology involves constructing three studies that compare QML models with traditional ML models; (i) \textit{Directional Classification} which predicts whether a stock’s price will rise or fall the next day (ii) \textit{Live trading} which tests how well a model decides when to buy or sell the share of stocks to maximize cumulative profit and (iii) \textit{Volatility forecasting} which estimates the size of price swings over a chosen horizon.
For each task, we compare QML models with classical counterparts that share the same inductive bias (to ensure fairness). Specifically, QNNs are evaluated against ANNs for Directional Classification, QLSTMs against LSTMs for live trading, and QSVRs against kernel SVRs for volatility forecasting. By aligning architectures and input representations, we isolate the additional representational power contributed by QML. Table \ref{tab:overview} summarizes the tasks, datasets, models, and evaluation criteria used for all the tasks considered in this paper.

\begin{table}[H]
  \centering
  \Large
  \caption{\scriptsize Overview of Forecasting Tasks, Datasets, Models, and Evaluation Metrics Used in our Benchmarking Study }\label{tab:overview}
  \begin{adjustbox}{max width=1.0\linewidth}
  \begin{tabularx}{\textwidth}{|>{\raggedright\arraybackslash}X
                               |>{\raggedright\arraybackslash}X
                               |>{\raggedright\arraybackslash}X
                               |>{\raggedright\arraybackslash}X|}
    \hline
    \textbf{Task} & \textbf{Dataset} & \textbf{Models Compared} & \textbf{Evaluation Metrics} \\
    \hline
    Directional classification & 5 Turkish equities / S\&P 500 / 5 US equities & QNN, ANN & Accuracy, AUC, Precision \\
    \hline
    Return-based live trading & S\&P 500 & QLSTM, LSTM & AUC, Annualized Return, Sharpe Ratio \\
    \hline
    Volatility forecasting & S\&P 500, AAPL, KCHOL, GARAN & QSVR, SVR, GARCH & QLIKE, Mean Squared Error \\
    \hline
  \end{tabularx}
  \end{adjustbox}
\end{table}

\subsection{Portfolio Selection} 
To ensure robustness and genuine predictive skill, we evaluate models across two distinct markets: the mature U.S. market and the emerging Turkish market, where deviations from the EMH are more likely. The U.S. universe combines the S\&P 500 benchmark with five representative firms spanning different capitalizations and sectors (Table~\ref{tab:us_composition}). A parallel Turkish universe is constructed to mirror this sectoral diversity, enabling assessment of model generalization across contrasting market regimes (Table~\ref{tab:tr_composition}).
Daily prices for each of the tickers (in both the markets) are retrieved via the \texttt{Yahoo Finance API}\footnote{\url{https://github.com/ranaroussi/yfinance}}. For individual equities, we use \emph{adjusted close} which accounts for stock splits and dividends. For the S\&P 500 benchmark, we use the close price directly. To align all tickers on a common trading calendar, we remove any day with missing observations, retaining only dates with complete data. With cleaned price series and finalized universes, we proceed to the forecasting tasks summarized in Table~\ref{tab:overview}.

\begin{table*}
  \centering
  \scriptsize
  \caption{\scriptsize Composition of the U.S.\ equity universe used across our studies}
  \label{tab:us_composition}
  \begin{tabularx}{\textwidth}{|c|c|X|c|X|}
    \hline
    \textbf{Role} & \textbf{Ticker} & \textbf{Company} & \textbf{Cap‐tier} & \textbf{Sector (GICS)} \\
    \hline
    \textbf{Benchmark} & \textbf{S\&P 500} & S\&P 500 & — & Broad market \\
    \hline
    1 & \textbf{AAPL} & Apple Inc. & Mega-cap & Information Technology \\
    \hline
    2 & \textbf{BA} & Boeing Co. & Large-cap & Industrials / Aerospace \& Defense \\
    \hline
    3 & \textbf{GILD} & Gilead Sciences Inc. & Mid-cap & Health Care / Biotechnology \\
    \hline
    4 & \textbf{DVN} & Devon Energy Corp. & Mid-cap & Energy / E\&P \\
    \hline
    5 & \textbf{LNC} & Lincoln National Corp. & Small-cap & Financials / Insurance \\
    \hline
  \end{tabularx}
\end{table*}

\begin{table*}
  \centering
  \scriptsize
  \caption{\scriptsize Composition of the Turkish equity universe used across our studies}
  \label{tab:tr_composition}
  \begin{tabularx}{\textwidth}{|c|c|X|c|X|}
    \hline
    \textbf{Role} & \textbf{Ticker} & \textbf{Company} & \textbf{Cap‐tier} & \textbf{Sector (GICS)} \\
    \hline
    1 & \textbf{KCHOL.IS} & Koç Holding A.Ş. & Large-cap & Industrials / Conglomerate \\
    \hline
    2 & \textbf{GARAN.IS} & Türkiye Garanti Bankası A.Ş. & Large-cap & Financials / Banking \\
    \hline
    3 & \textbf{TUPRS.IS} & Türkiye Petrol Rafinerileri A.Ş. & Mid-cap & Energy / Refining \\
    \hline
    4 & \textbf{ULKER.IS} & Ülker Bisküvi Sanayi A.Ş. & Small-cap & Consumer Staples \\
    \hline
    5 & \textbf{TCELL.IS} & Türkcell İletişim Hizmetleri A.Ş. & Mid-cap & Communication Services \\
    \hline
  \end{tabularx}
\end{table*}

\vspace{1em}


\subsection{Directional Classification}\label{sec:Directional Classification}

Building on the task definition outlined in Section~\ref{sec:Background}, we now apply directional classification to our portfolio universes. Each day is labeled as “up” or “down” according to the binary rule in Eq.~\ref{eq:binary_target}.  Both QML and classical ML models then learn to predict these labels using a diverse collection of rich features, the details of which we describe below.

\subsubsection{Feature Selection}
\begin{figure*}
\centering
\includegraphics[width=0.80\textwidth]{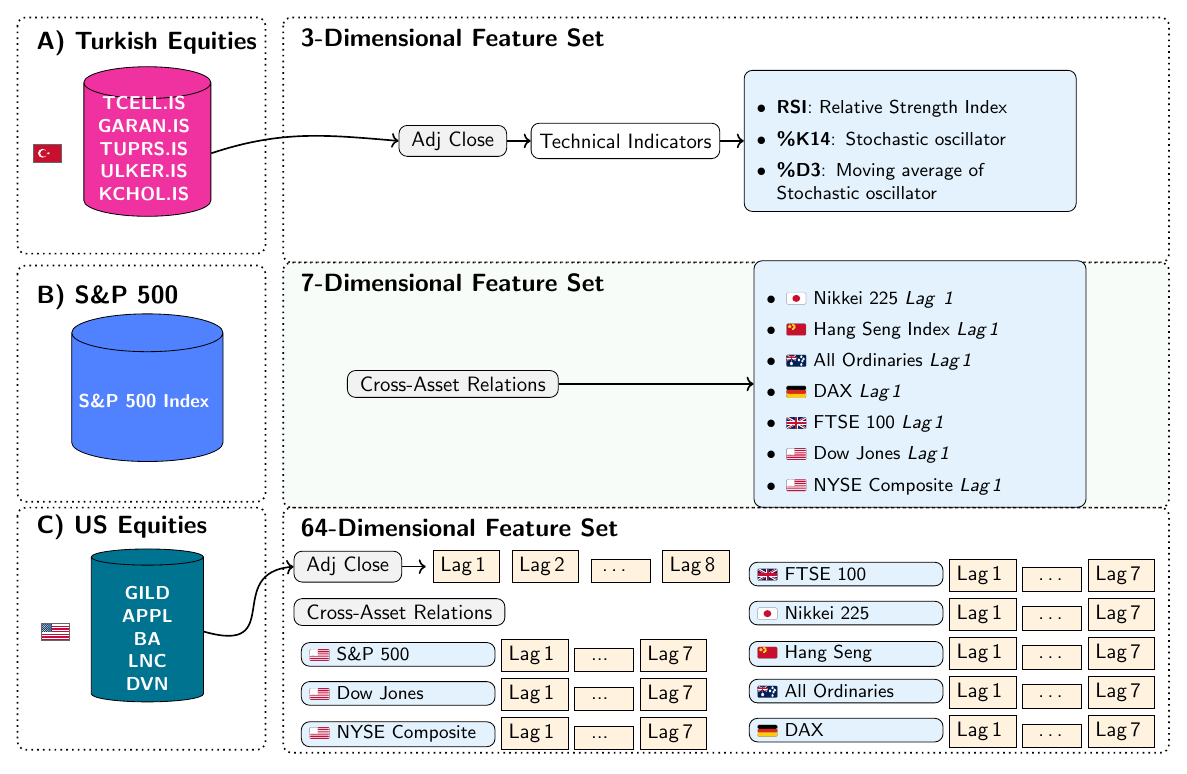}
\caption{\scriptsize \scriptsize
We use 3 feature regimes in our first study.  \textit{Adj Close} refers to stock returns computed from adjusted‑close prices, which account for splits and dividends. \textit{Lag} indicates the previous day's return of the stock. (A) For Turkish equities, a 3-D feature set of technical indicators is constructed from adjusted close prices. (B) For S\&P 500 index, a 7-D feature set is derived from cross-asset relations.  (C) For selected US equities, a 64-D feature set includes the past 8 day returns of the equity itself as well as the 7 day returns of eight major indices. 
}
\label{fig:feature_regimes}
\end{figure*}

Effective feature selection is central to extracting meaningful signals from noisy price data. We consider three feature regimes for directional classification (Fig.~\ref{fig:feature_regimes}), which illustrate how the feature sets are constructed for Turkish equities (3-D), the S\&P 500 index (7-D), and selected U.S. equities (64-D). For each feature set, we utilize three main types of features: (1) \textbf{Technical indicators} that are derived from historical prices and capture patterns like momentum and trend. Frequently used indicators across our studies include moving averages (MA), the relative strength index (RSI), and the moving average convergence/divergence (MACD) oscillator, (2) \textbf{Lagged (self‑)returns} that capture short‑term autocorrelation in the target stock’s own history, and (3) \textbf{Cross-asset lagged returns} that capture how past returns of related stocks or indices influence the target stock. Since many assets tend to react similarly to broad market or economic events, their prices often rise and fall together.
By combining these three sources, we build feature sets that are rich enough to exploit both temporal dependencies and cross‑sectional relationships when forecasting the next‑day return direction.


\paragraph{Low-Dimensional Feature Set}\label{para:lowdim}\mbox{} As illustrated in Fig.~\ref{fig:feature_regimes}(A), we define a \emph{3-dimensional} feature set containing three input variables ($d = 3$). 
For this feature set we convert the adjusted‑close prices of Turkish equities (Table~\ref{tab:tr_composition}) into three momentum indicators: (1) \textbf{RSI-14} The 14-day Relative Strength Index (RSI-14) measures the magnitude of recent gains versus losses on a 0--100 scale, (2)\textbf{\%K-14} The raw stochastic oscillator over 14 days positions today’s price relative to the 14-day high-low range (\%K14), and (3) \textbf{Moving Average of \%K14} a 3-day simple moving average of \%K14 smooths out noise and confirms true momentum shifts.
This 3-dimensional feature set, consisting of various momentum indicators, captures whether a stock has been overbought or oversold in the past 14 trading days.

\vspace{1em}


\paragraph{Medium-Dimensional Feature Set}

Fig.~\ref{fig:feature_regimes}(B) depicts the \emph{7-dimensional} feature set which contains seven input variables ($d = 7$) aimed at predicting the direction of returns for the S\&P 500 (Table~\ref{tab:us_composition}). In this feature set, we take the viewpoint that markets are interconnected \cite{raddant2021interconnectedness}. What happens in other regions often foreshadows U.S. trading. Usually, indices from the Asia-Pacific and European markets are known before the U.S. cash close (21:00 UTC) \cite{jiao2017predicting}. Hence, we test whether those “early” global returns contain exploitable information to predict the closing direction of the S\&P 500 for QNNs. Our feature set, therefore, consists of same-day log-returns of five non-U.S. indices and two U.S. proxies (DJI, NYA). Table~\ref{tab:utc_hours} provides the regional breakdown and time-zone alignment of all indices used.

\begin{table}[h]
\centering
\caption{\scriptsize \scriptsize Indicative UTC trading hours for each global index in the 7-dimensional feature set. We adopt a close-to-close protocol: APAC/EU closes occur before the U.S. cash close (21:00 UTC),  and U.S. proxies are \emph{lagged by one day} (we use $\text{DJI}_{t-1}$ and $\text{NYA}_{t-1}$ when predicting day-$t$ S\&P\,500). This guarantees every input feature is drawn solely from information available prior to prediction.}
\Large
\begin{adjustbox}{max width=1.0\linewidth}
\begin{tabularx}{\textwidth}{|X|X|X|}

\hline
\textbf{Index} & \textbf{Region} & \textbf{UTC Hours} \\
\hline
Nikkei 225 (N225) & Japan & 00:00 – 06:00 \\
Hang Seng (HSI) & Hong Kong & 01:30 – 08:00 \\
All Ordinaries (AORD) & Australia & 00:00 – 06:00 \\
DAX (GDAXI) & Germany & 08:00 – 16:30 \\
FTSE 100 (FTSE) & United Kingdom & 08:00 – 16:30 \\
Dow Jones (DJI) & United States & 14:30 – 21:00 \\
NYSE Composite (NYA) & United States & 14:30 – 21:00 \\
\hline
\end{tabularx}
\end{adjustbox}
\label{tab:utc_hours}
\end{table}


\paragraph{High-Dimensional Feature Set}

In Fig.~\ref{fig:feature_regimes} (C), we summarize the \emph{64-dimensional} feature set ($d = 64$) used for U.S. equities (Table~\ref{tab:us_composition}). This regime provides models with a deeper temporal and cross-sectional window, combining information from both the stock’s own history and major global indices. Specifically, we construct a lag-matrix embedding that includes seven days of returns from eight global indices (56 features) together with the stock’s own past eight daily returns.  The resulting 64 features are then fed to both QNNs and ANNs.

\subsubsection{Feature Pre-processing} 
\begin{figure}[h]
    \centering
    \includegraphics[width=1.0\linewidth]{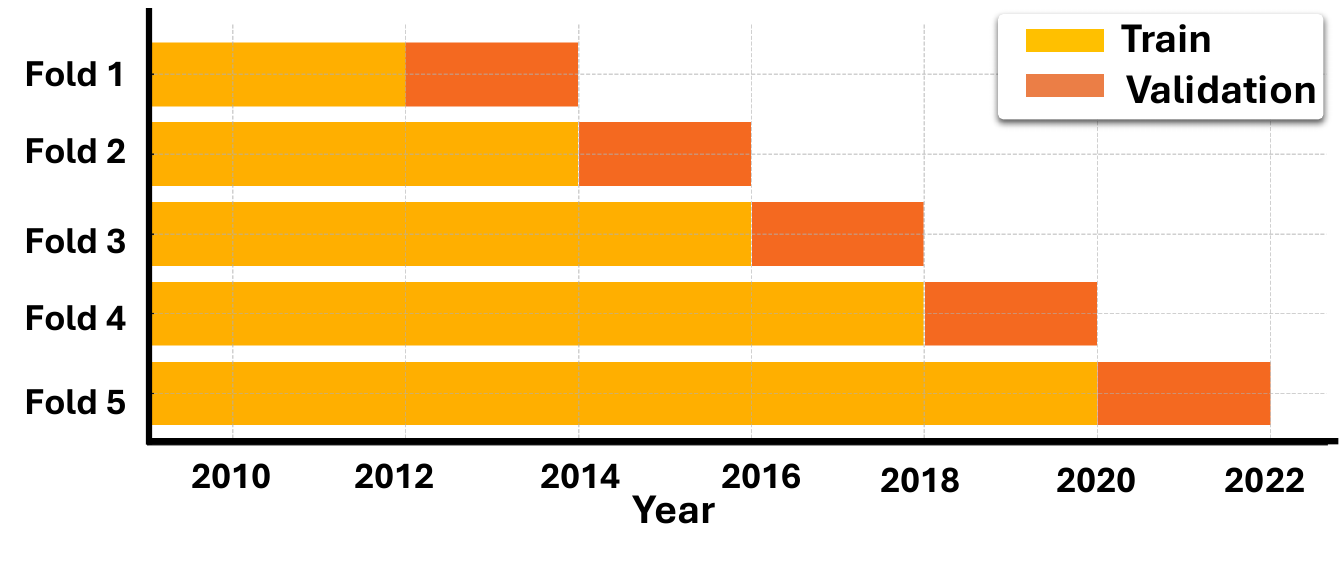}
    \caption{\scriptsize Walk-forward cross-validation splits used across five folds for directional classification. Each fold uses an expanding training window followed by a fixed validation window.}
    \label{fig:cv_splits}
\end{figure}

For each of these feature sets (low, medium, and high dimensional), the raw input features are scaled to the range $[0,1]$ using \texttt{min-max normalization}. This prevents any single feature from dominating the model decision and ensures smooth optimization. The min-max scaler is fitted on the training fold and applied to the corresponding test fold to prevent look-ahead bias. Exceptions arise where feature-specific constraints, such as bounded momentum indicators, make scaling unnecessary (e.g., the low-dimensional Turkish ANN baseline). For each feature set, we adopt a walk-forward, expanding-window cross-validation scheme with five folds. The train-validation splits are presented in Fig.~\ref{fig:cv_splits}. We report results for the final fold, which has a training period from \textit{2009-01-07} to \textit{2019-12-31} and a testing period from \textit{2020-01-02} to \textit{2021-12-31}.

\subsubsection{Models used for Directional Classification}

Since there is no universally optimal architecture in either the classical or quantum setting, we conduct an exhaustive model search for both QNNs and ANNs. The search ensures that comparisons reflect differences in quantum versus classical structure, rather than disparities in model capacity.

\paragraph{QNN Variants.}

\begin{figure*}
    \centering
    \includegraphics[width=1\textwidth]{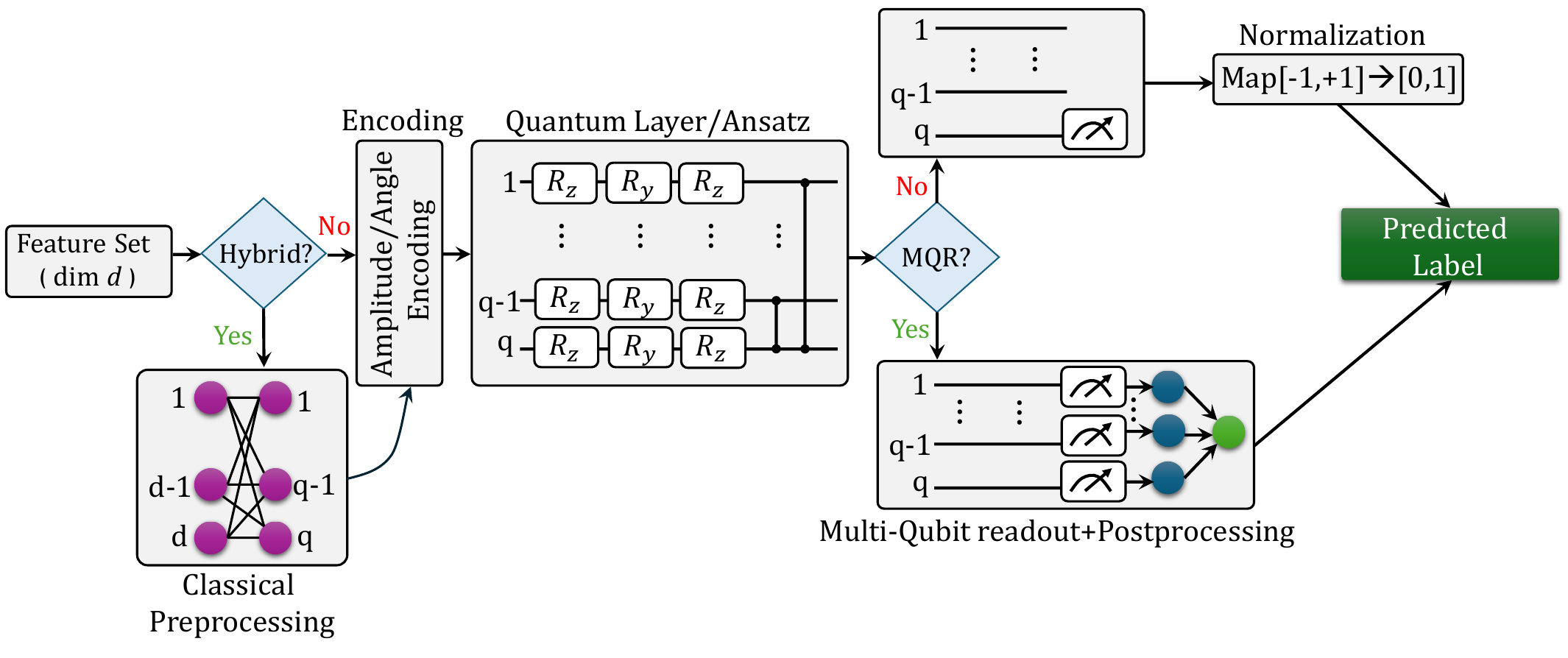}
    \caption{\scriptsize \textbf{QNN architectures explored for directional classification}.  
    A classical feature set of dimension $d$ is fed into the model. If $\texttt{Hybrid} = \texttt{True}$, the QNN optionally preprocesses these features using a classical layer that maps the $d$ inputs to $q$ outputs. The resulting $q$-dimensional feature vector is then encoded into a register of $q$ qubits via \emph{angle} or \emph{amplitude} encoding. A parameterised ansatz of rotation gates ($R_x$, $R_y$, $R_z$) and entangling operations is applied. After the circuit, either (1) a \emph{single-qubit readout} measures one qubit and the expectation value is normalised from $[-1, +1]$ to $[0, 1]$, or (2) a \emph{multi-qubit readout} measures all $q$ qubits and aggregates the results. The post-processed measurement outcome yields the final predicted label.
}
    \label{fig:qnn_ann_arch}
\end{figure*}

We explore four QNN families along two design axes: (1) the presence or absence of a \textit{classical preprocessing layer} (\texttt{Hybrid = True}), which maps $\mathbb{R}^d \to \mathbb{R}^q$ before quantum encoding, and (2) the readout strategy. The overall illustration is shown in Fig. \ref{fig:qnn_ann_arch}. In the \textbf{single-qubit readout} case (\texttt{MQR = False}), the expectation value of a designated qubit is measured and simply rescaled from $[-1, +1]$ to $[0, 1]$; \emph{no classical postprocessing layer is applied}. In the \textbf{multi-qubit readout} case (\texttt{MQR = True}), all $q$ qubits are measured and their outcomes are aggregated through a classical dense layer to yield the final prediction. This yields four architectural variants:

\begin{itemize}
    \item \textbf{QNN-SQ}: fully quantum, i.e., no classical pre and pot processing layers, single-qubit readout.  
    \item \textbf{QNN-MQ}: same as QNN-SQ but multi-qubit readout with classical postprocessing.  
    \item \textbf{Hybrid-SQ}: classical input projection + single-qubit readout.  
    \item \textbf{Hybrid-MQ}: classical input projection + multi-qubit readout and classical post-processing layer.  
\end{itemize}

For each QNN variant, we conduct a grid search over the depth of the quantum layer ($L \in \{1,\dots,6\}$) and the readout strategy. The choice of encoding depends on feature dimensionality. For the 3-D and 7-D feature sets, we employ angle encoding, where each feature requires a dedicated qubit. This results in 3 and 7 qubits, respectively, both feasible within current NISQ limits. In contrast, applying angle encoding to the 64-D feature set would require 64 qubits, which is impractical in the NISQ era. Instead, we use amplitude encoding, which embeds $2^n$ features into $n$ qubits, reducing the requirement to $n = 6$ qubits for 64 features. Since the optimal qubit count for amplitude encoding is not known beforehand, we treat $n \in \{2,\dots,6\}$ as a hyperparameter in the grid search for the 64-D case, while the qubit numbers for the 3-D and 7-D cases remain fixed.

\paragraph{ANN Baselines}
To benchmark against classical networks, we deliberately employ shallow ANNs (1–2 hidden layers) consistent with prior works \cite{jiao2017predicting, zhong2017forecasting}. For each feature regime, we fix the input dimension $d$, perform a grid search over hidden-layer widths, and select the optimal ANN based on validation AUC. The chosen baselines are:  
\begin{itemize}
    \item \textbf{Low-dimensional (3-D)}: [3–11–1] with one hidden layer of 11 neurons, totaling 56 parameters.  
    \item \textbf{Medium-dimensional (7-D)}: [7–32–16–1] with two hidden layers (32 and 16 neurons), totaling 801 parameters.  
    \item \textbf{High-dimensional (64-D)}: [64–32–1] with one hidden layer of 32 neurons, totaling 2,113 parameters.  
\end{itemize}
This ensures QNNs and ANNs have comparable parameter counts: $\sim$40–60 (low), $\sim$100–1000 (medium), and $\sim$2000–2200 (high).

\paragraph{Selection Criterion}\label{para:selection_criterion}
Across both QNN and ANN families, candidate models are trained on expanding windows and evaluated via mean AUC across five validation folds. The architecture with the highest mean AUC is selected; if multiple architectures achieve the same AUC, the one with fewer parameters is chosen. Thus, Hybrid QNNs and purely classical ANNs are both optimized via search, but they differ fundamentally in how capacity is distributed: QNNs balance quantum encoding and variational depth with optional classical components, whereas ANNs rely solely on hidden-layer depth and width.

\subsubsection{Evaluation Metrics for Directional Classification}

To assess predictive performance in the directional classification task, we use metrics that extend beyond directional accuracy (Acc). While Acc provides a baseline, it can be misleading under class imbalance—common in financial returns, where up and down days are rarely equal. For instance, if 55\% of days are positive, a naïve “always up” model achieves 55\% Acc without capturing meaningful signal. To address this, we complement accuracy with threshold-independent and class-sensitive metrics.

\begin{itemize}
    \item \textbf{Accuracy (Acc):} Proportion of correct predictions.
    
    \item \textbf{Precision (P):} Among all days where the model predicted an upward movement, precision measures how many actually resulted in positive returns. It gauges the reliability of a model’s buy signals:
    \[
    \text{Precision} = \frac{\text{True Positives}}{\text{True Positives} + \text{False Positives}}
    \]
    
    \item \textbf{Recall (R):} Among all days with actual positive returns, recall measures how many were successfully identified by the model. It reflects the model's sensitivity to upward movements:
    \[
    \text{Recall} = \frac{\text{True Positives}}{\text{True Positives} + \text{False Negatives}}
    \]
    
    \item \textbf{Area Under the ROC Curve (AUC):} AUC assesses the model’s ability to rank positive instances higher than negative ones across all classification thresholds. Unlike accuracy, AUC is threshold-agnostic and robust to class imbalance. It is especially informative in cases where the proportion of up/down days deviates from 50\%.
\end{itemize}

\begin{table*}[b]
\centering
\footnotesize
\renewcommand{\arraystretch}{1.25}
\caption{Statistical characteristics of each market regime, including annualized return ($\mu_{\rm ann}$), volatility ($\sigma_{\rm ann}$), skewness, kurtosis, maximum drawdown (Max DD), and the Sharpe ratio. Computed on the full sample for each period, these statistics highlight the distinct profiles of each regime. For brevity, folds are labeled as F1–F4: F1 = Global Financial Crisis (2008–2009), F2 = Pre-Covid (2018–2019), F3 = Covid Shock \& Recovery (2020–2021), and F4 = Post-pandemic (2022–2024).}
\label{tab:regimestats}
\begin{tabular*}{\textwidth}{@{\extracolsep{\fill}} c c c 
S[table-format=-2.1] 
S[table-format=2.1]  
S[table-format=-1.2] 
S[table-format=2.2]  
S[table-format=-2.1] 
S[table-format=-1.2] 
@{}}
\toprule
\textbf{Fold} & \textbf{Start} & \textbf{End} &
{$\mu_{\rm ann}$ (\%)} & {$\sigma_{\rm ann}$ (\%)} &
\textbf{Skew} & \textbf{Kurtosis} & \textbf{Max DD (\%)} & \textbf{Sharpe} \\
\midrule
F1 & 2008-01-02 & 2009-12-31 & -13.7 & 34.9 & -0.11 & 7.30 & -53.3 & -0.39 \\
F2 & 2018-01-02 & 2019-12-31 &   9.5 & 15.0 & -0.62 & 6.64 & -19.8 &  0.63 \\
F3 & 2020-01-02 & 2021-12-31 &  19.4 & 26.2 & -1.05 & 17.76& -33.9 &  0.74 \\
F4 & 2022-01-03 & 2024-12-31 &   7.0 & 17.5 & -0.22 & 4.79 & -25.4 &  0.40 \\
\bottomrule
\end{tabular*}
\end{table*}

\subsection{Live Trading}

While predicting return direction is informative, it is the translation of forecasts into trading signals that ultimately determines long-term market success. In our second study, we embed both QLSTM and classical LSTM models within a backtesting framework to compare their ability to deliver risk-adjusted returns. The evaluation is conducted on the S\&P 500 index across four historical market regimes and assessed using trading metrics such as annualized return, volatility, and Sharpe ratio. The following sections describe the market regimes and feature sets used for live trading, followed by the model architectures and evaluation protocols applied to assess the performance of each model.

\subsubsection{Market regime for S\&P 500} \label{sec:sp500_regimes}

We evaluate live trading on the S\&P 500 from 2008–2024 which we segment into four representative regimes: (i) the \textit{Global Financial Crisis} (2008–2009) (ii) the \textit{Pre-Covid era} (2018–2019),  (iii) the \textit{Covid Shock \& Recovery} (2020–2021) and (iv) the \textit{Post-pandemic era} (2022–2024). These regimes differ substantially in terms of average returns, volatility, and distributional shape, as summarized in Table~\ref{tab:regimestats}.

To validate these regimes, we conduct three statistical tests that compare the return patterns of each of them. The \textit{mean-difference test} ($t$-test) \cite{Student1908} checks whether the average returns in two regimes are significantly different. The \textit{variance test} (Brown–Forsythe test) \cite{BrownForsythe1974} looks at how volatile the returns are in each regime. Finally, the \textit{distribution shape test} 
(Kolmogorov–Smirnov)\cite{Welch1947} compares the full return distributions to see if the overall behavior, including outliers and asymmetries, changes from one regime to another. The results of these distributional tests are shown in Table~\ref{tab:distributiontests}:

\begin{table}[h]
\centering
\caption{\scriptsize %
Statistical test results comparing return distributions across S\&P 500 market regimes. 
Each row shows the $t$-statistic and $p$-value for the difference in means ($\Delta\mu$), 
the Brown–Forsythe $F$-statistic for differences in variances ($\Delta\sigma$), 
and the Kolmogorov–Smirnov (KS) distance for comparing entire distribution shapes. 
Significant $p$-values (e.g., $p < 0.001$) indicate meaningful differences across regimes. F1-F4 are regime labels.
}
\begin{adjustbox}{max width=1.0\linewidth}
\Large
\begin{tabular}{|c|c|c|c|}
\hline
\textbf{Comparison} & $\Delta\mu$ $t$-stat (p-val) & $\Delta\sigma$ $F$-stat (p-val) & KS $D$ (p-val) \\
\hline
F1 vs F3 & $-1.08$ (0.283) & 29.37 ($7.48 \times 10^{-8}$) & 0.14 ($9.03 \times 10^{-5}$) \\
F1 vs F2 & $-0.87$ (0.387) & 113.20 ($3.98 \times 10^{-25}$) & 0.19 ($2.09 \times 10^{-8}$) \\
F2 vs F4 & $0.17$ (0.868) & 14.98 ($1.14 \times 10^{-4}$) & 0.08 ($3.03 \times 10^{-2}$) \\
\hline
\end{tabular}
\end{adjustbox}

\label{tab:distributiontests}
\end{table}

The $t$-tests indicate that average returns do not significantly differ between regimes (all $p$-values are greater than 0.05). This suggests that long-term mean performance may be similar. However, the variance tests show highly significant differences in return volatility across all comparisons with $p$-values below 0.001. This means the level of risk fluctuated sharply across regimes. In addition, the KS test results confirm that the overall shape of return distributions changes meaningfully. These results validate the separation of these periods into distinct market regimes. 

Overall, by stress-testing both QLSTM and LSTM across the distinct market regimes of S\&P 500, we ensure that any performance gain is robust and not attributable to a single market condition. 

\subsubsection{Feature Selection and Preprocessing}

Since our live trading models are recurrent (see the next section), we select the features with temporal dependencies that both classical and quantum networks can exploit. Our feature set integrates short- and medium-term market signals derived from adjusted-close prices of the S\&P 500 index. The feature set consists of: (1) the \textbf{previous-day log return}, $r_t = \ln(P_t / P_{t-1})$, which captures daily price changes and reflects short-term fluctuations, (2) the \textbf{MACD line}, defined as the difference between the 12-day and 26-day exponential moving averages (EMAs) \cite{appel2005}, which tracks underlying market momentum, (3) the \textbf{MACD signal}, a 9-day EMA of the MACD line that smooths medium-term trend shifts, and (4) the \textbf{RSI-14}, a 14-day Relative Strength Index which quantifies the balance of recent gains and losses. Collectively, this four-dimensional feature set captures both short-term return dynamics and medium-horizon momentum behavior, making it well-suited for recurrent neural architectures.

We stack these features into rolling windows of length $T = 10$ trading days. Each input sample is thus a matrix $\mathbf{x}_t \in \mathbb{R}^{10 \times 4}$ where each row corresponds to four indicators on a given day. This window length strikes a balance between capturing recent momentum and keeping the recurrent circuit depth feasible for quantum simulations. 

\begin{figure}[H]
\centering
\includegraphics[width=1.0\linewidth]{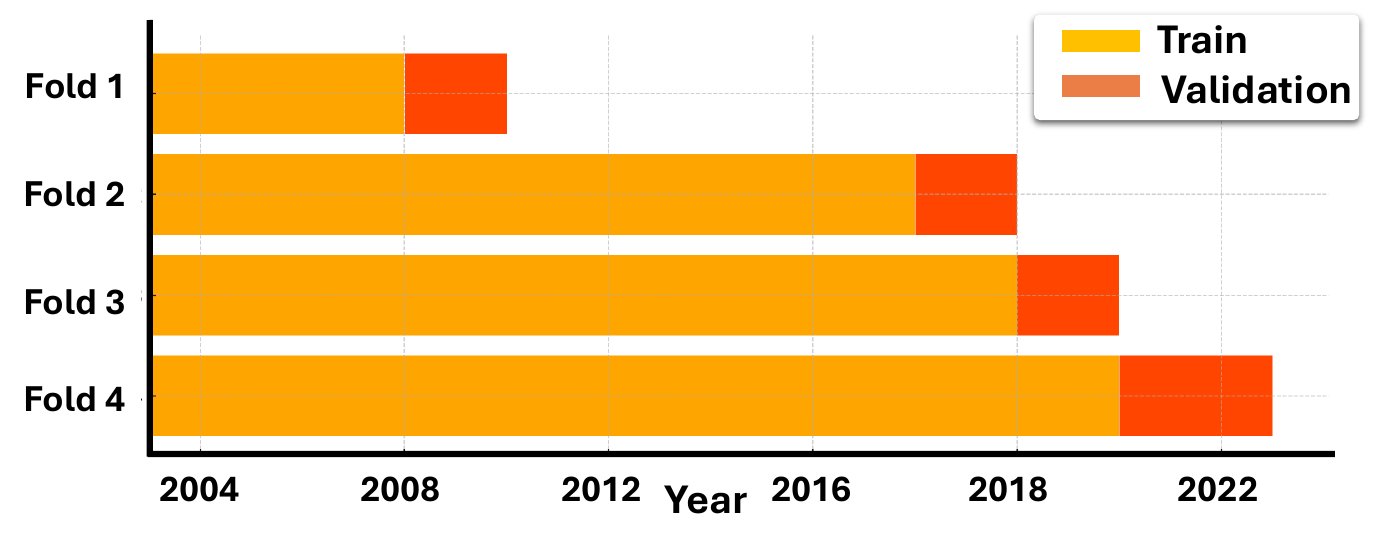}
\caption{\scriptsize %
Walk-forward cross-validation splits are used for live trading across four market regimes in the S\&P 500. 
Each regime corresponds to a historical period identified in Table~\ref{tab:regimestats}, and is divided into 
train, early-stop, model-select, and threshold-calibration slices. This structure ensures out-of-sample 
integrity and mimics a realistic trading pipeline.%
}
\label{fig:walkforward_study2}
\end{figure}

All features are normalized to the $[0, 1]$ range using \texttt{min-max scaling}. Similar to the directional classification, the scaler is fitted only on the training portion of the walk-forward split to avoid lookahead bias. We use an expanding window cross-validation strategy with four folds as shown in Fig.~\ref{fig:walkforward_study2}. 

\subsubsection{Models used for Live Trading}

In this section, we outline the model architectures (classical and hybrid) used for return-based live trading. Since no recurrent architecture is universally optimal, we compare quantum and classical LSTMs under matched parameter budgets to isolate the effect of quantum gating.

\begin{figure}[H]
    \centering
    \includegraphics[width=\linewidth]{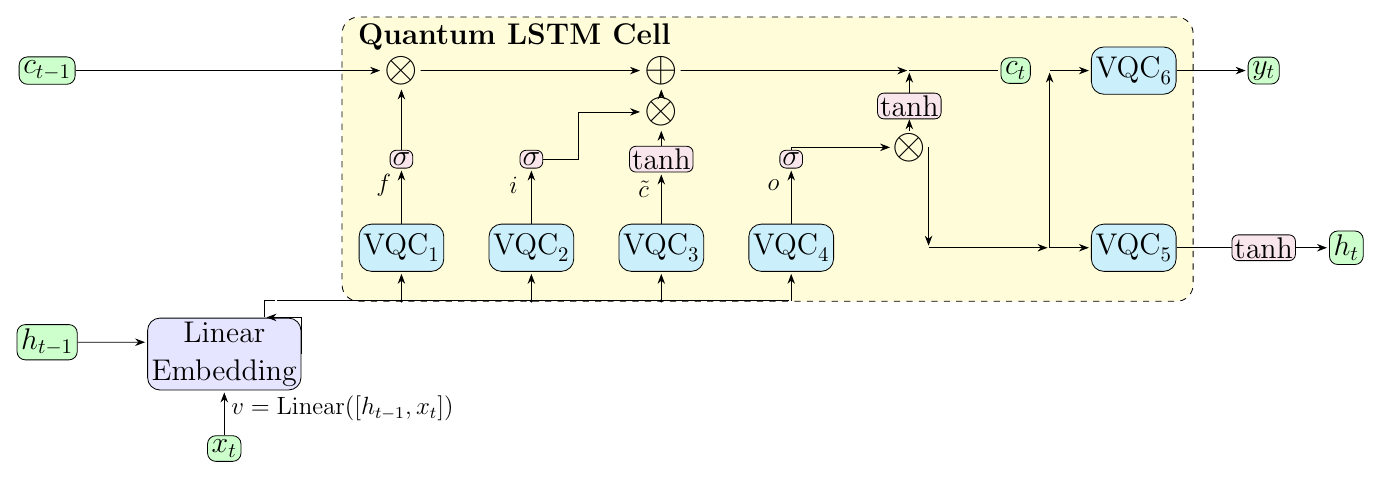}
    \caption{\scriptsize \textbf{QLSTM architecture used in live trading.} 
    A classical linear embedding combines the previous hidden state and current input before sending them into VQCs. 
    Each gate of the LSTM is modeled by its own VQC, which replaces the usual linear transformations in classical LSTM. Additional VQCs produce the hidden state and final output, which makes the design a direct quantum analog of the classical LSTM.}
    \label{fig:qlstm_cell}
\end{figure}

\paragraph{QLSTM variants}  Our QLSTM design builds on the architecture proposed by \cite{chen2022quantum}, where each classical gate in a standard LSTM cell is replaced with a variational quantum circuit (VQC). The computational flow of the long short-term memory is depicted in Fig.~\ref{fig:qlstm_cell}.

 Analogous to the QNN search, we vary two design axes for our QLSTM: (i) Qubit width ($q \in {3,4,5}$), which sets the embedding dimension for each gate, and (ii) Circuit depth ($L \in {2,3,4,5,6}$), which controls the expressivity of the variational ansatz. Each VQC block employs a hybrid input layer that maps classical inputs into qubit rotations followed by a multi-qubit $Z$-basis readout to produce features (see Hybrid-MQ in our directional classification study). Together, these design choices define a family of QLSTM models with parameter counts in the $10^2$–$10^3$ range.
 
\paragraph{LSTM variants} As a baseline, we construct LSTM models with hidden state sizes $h \in {3, 4,5}$ and stacked layers $L \in {2,3,4,5,6}$. The gating functions are implemented using linear weights $(W_f, W_i, W_c, W_o)$. The resulting parameter counts match the QLSTM family, which ensures that differences reflect quantum versus classical computation rather than raw size.

\paragraph{Training Objective} The output from both models is compared against the actual labels using a weighted BCE loss $\mathcal{L}_{\rm wbce}$ to account for the class imbalance in directional labels. These model outputs serve as our starting point for implementing our backtesting strategy.

\subsubsection{Backtesting Strategy}

Backtesting involves simulating a trading strategy on historical data to see how it would have performed in live deployment. In our setup, we implement a threshold-based rule to convert model outputs into discrete trading signals.
For each trading day $t$, the model produces a probability $\hat{y}t \in [0,1]$ of a positive next-day return $r{t+1}$. This probability is converted into a position signal $s_t \in \{-1,0,+1\}$. The position signal $+1$ denotes a long position while $-1$ denotes a short position. On the other hand, $0$ reflects a neutral stance in which the strategy abstains from taking any position.

To map the model output $\hat{y}_t$ into a position signal $s_t$, we define two thresholds, $\tau_{\text{long}}$ and $\tau_{\text{short}}$. When $\hat{y}_t > \tau_{\text{long}}$, the probability is interpreted as strong evidence of a positive return, prompting a long signal. When $\hat{y}_t < \tau_{\text{short}}$, it is treated as strong evidence of a negative return, prompting a short signal. Values that lie between the two thresholds fall into a no-trade region, reflecting cases where the model’s prediction lacks sufficient confidence to justify taking a position. Formally:

\begin{equation}
s_t = 
\begin{cases}
+1 & \text{if } \hat{y}_t > \tau_{\text{long}} \quad \text{(Go Long)}, \\
-1 & \text{if } \hat{y}_t < \tau_{\text{short}} \quad \text{(Go Short)}, \\
\;\;0 & \text{otherwise (No Trade)}.
\end{cases} 
\end{equation}

A 0.05\% transaction fee is applied whenever the position changes, ensuring profitability is evaluated net of trading costs. Portfolio equity evolves according to:

\begin{equation}
\text{Equity}_{t+1} = \text{Equity}_t(1 + s_t \cdot r_{t+1} - \text{Cost}_{t+1}),  
\end{equation}

where $\text{Equity}_t$ is portfolio value at the start of day $t$, and $\text{Cost}_{t+1}$ reflects trading frictions. The term $s_t \cdot r_{t+1}$ captures profits or losses from the chosen position.

As a benchmark, we compare the generated signals from both QLSTM and LSTM against a simple buy-and-hold strategy. Under buy-and-hold, an investor goes long $s_{t}=+1$ on the S\&P 500 at the start of each fold and holds the position throughout the entire period. This strategy ignores any predictive signal and assumes the market has a positive drift over time. Returns under buy-and-hold are computed by compounding daily log-returns. 

\subsubsection{Validation Scheme}

\begin{figure}[H]
\centering
\includegraphics[width=1.0\linewidth]{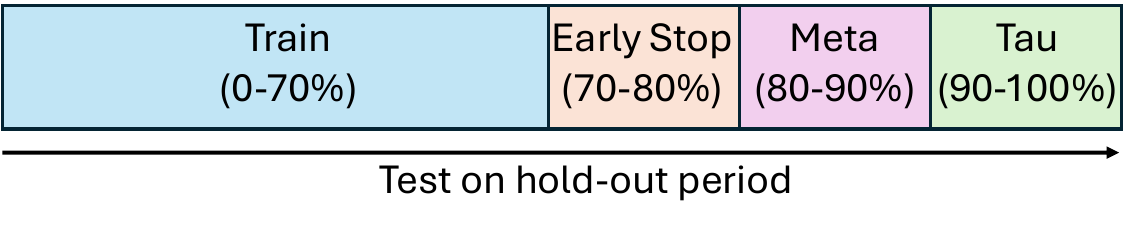}
\caption{\scriptsize %
Illustration of the 70/10/10/10 split applied to each market regime fold in the live trading experiment. 
Each phase corresponds to a distinct role: (1) model training, (2) early stopping via ES AUC, 
(3) model selection based on AUC, and (4) threshold calibration to maximize the Sharpe ratio. 
This design ensures strict out-of-sample validation and mimics a real-world backtesting workflow.%
}
\label{fig:regime_splits}
\end{figure}

Each of the four market regime folds is split in the ratio 70/10/10/10 (see Fig.~\ref{fig:regime_splits}) for training, early stopping, model selection, and threshold calibration phases. The first 70\% of each data regime is used to optimize weights. The subsequent 10\% of each regime serves as an \emph{early-stopping slice} and the training is stopped when the Early-Stop AUC (ES AUC), which is defined as the standard ROC-AUC, fails to improve for a fixed patience window (30 epochs). The next 10\% of the data compares different model configurations under a leak-free setting and picks the one with the highest AUC. 
Finally, the last 10\% find the optimal thresholds of $(\tau_{\text{long}}, \tau_{\text{short}})$. Both thresholds are swept within a grid:
    \[
    \tau_{\text{long}} \in [0.5,0.9], \quad \tau_{\text{short}} \in [0.1,0.5].
    \] 
For each threshold pair, we simulate trading using the derived signal and compute the \textbf{Sharpe Ratio}. The pair that maximizes Sharpe is selected for final backtesting. By separating model training, selection, and threshold calibration, we ensure that the final backtest is leak-free. 
Finally, we point out that since our folds maximize the Sharpe ratio during backtesting, 
our evaluation focuses on \textit{risk-adjusted profitability} and not just predictive accuracy. This distinction is critical since a model may achieve high AUC but still perform poorly in trading if its predictions do not align well with profitable return magnitudes.

All final models are selected based on the configuration that maximizes Sharpe on the hold-out $\tau$-slice (the 10\% calibration split). This ensures that model choice is aligned with economic performance rather than purely statistical accuracy.

\subsubsection{Evaluation Metrics}

While AUC remains useful for measuring classifier ranking ability, it does not guarantee that forecasts translate into profitable trades. Once model outputs are converted into trading signals, economic metrics become more relevant. We therefore evaluate strategies using the following measures:

\begin{itemize}
    \item \textbf{Annualized Return (ARC):} Measures the geometric average yearly return. 
    Let $r_t$ be the per‐period return, $T$ the total number of periods, and $N_{\rm trading}$ the number of trading periods per year. Defining $R_{\rm tot} = \prod_{t=1}^T (1 + r_t)$, ARC is
    \[
    \mathrm{ARC} = R_{\rm tot}^{\tfrac{N_{\rm trading}}{T}} - 1.
    \]

    \item \textbf{Annualized Standard Deviation (ASD):} Captures the volatility of returns, reflecting the typical size of annual swings. With $\sigma_r$ denoting return volatility,
    \[
    \mathrm{ASD} = \sigma_r \sqrt{N_{\rm trading}}.
    \]

    \item \textbf{Sharpe Ratio:} Evaluates excess return per unit of total risk and is standard in portfolio optimization. With $r_f$ as the per-period risk-free rate,
    \[
    \mathrm{Sharpe} = \frac{\mathrm{ARC} - r_f}{\mathrm{ASD}}.
    \]

    \item \textbf{Sortino Ratio:} A downside-focused analogue of Sharpe that penalizes only negative volatility. With $\sigma_d$ denoting downside deviation,
    \[
    \mathrm{Sortino} = \frac{\mathrm{ARC} - r_f}{\sigma_d}.
    \]
\end{itemize}

By calibrating thresholds on a hold-out slice to maximize Sharpe, we ensure that model evaluation emphasizes risk-adjusted profitability rather than directional accuracy alone.

\subsection{Volatility Forecasting}

Volatility plays a critical role in financial markets as a proxy for risk. 
In this study, we focus on predicting short-term realized variance as our volatility target. We benchmark QSVR against classical SVR and a standard econometric baseline, GARCH(1,1). This setup allows us to evaluate whether quantum kernels provide a meaningful advantage in capturing volatility dynamics beyond established statistical and ML methods.

Our prediction target is the 5-day rolling realized variance:
\begin{equation}  
RV_t  = \frac{1}{5} \sum_{i=0}^4 r_{t-i}^2.  
\end{equation}

where $RV_t$ denotes the realized variance at time $t$.
Thus, the supervised learning target is the one-step-ahead realized variance:
\begin{equation}  
y_{t+1} = RV_{t+1}.  
\end{equation}

We evaluate forecasts on two U.S. assets (S\&P 500 index, Apple Inc.) and two Turkish assets (Koç Holding A.Ş.\ and Türkiye Garanti Bankası A.Ş.).

\subsubsection{Feature Selection and Preprocessing}

For each asset, we construct a feature vector that combines past returns and past realized variance of that asset. Our choice of features follows previous works \cite{faldzinski2020forecasting} \cite{zhuo2024hybrid}, which show the effectiveness of these features in modeling volatility dynamics. The past realized variance captures recent trends in volatility, while past returns capture the most recent price movements. 
Let $r_t$ be the past returns and $RV_t$ be the realized variance at time $t$. Our feature set is composed of the past $p$ values of returns and past $q$ values of realized variance:
\begin{equation}
x_t = [r_{t:t-p+1},\; RV_{t:t-q+1}] \in \mathbb{R}^{p+q}.    
\end{equation}

For training, we adopt an expanding-window forecasting approach for each equity ticker. We begin with an initial training set of 720 days, and every 120 days, the model is retrained using all available past data to account for potential regime shifts. To examine robustness across different sample lengths, we also experiment with total training window lengths $L \in \{1000,\ 2000,\ 3000,\ 4000\}$ days.

\subsubsection{Models used for Volatility Forecasting}

In order to explore whether Quantum Kernels can represent certain high-order correlations more efficiently than classical kernels, we compare kernel-based SVR methods against QSVRs. As a point of comparison, we include the GARCH(1,1) model, which is a standard econometric approach for volatility forecasting.  We outline its specification below to establish the benchmark against which kernel-based and quantum-enhanced models are evaluated.

\paragraph{Garch(1,1)}

GARCH(1,1) captures the idea that today’s volatility depends both on yesterday’s market shock (the deviation of the return $r_t$ from its mean $\mu$) and on yesterday’s level of volatility. As a result, under this formulation, the conditional variance $\sigma_{t+1}^2$ of the return at time $t+1$ is expressed as a linear function of the previous day’s squared shock and the previous day’s variance estimate:

\begin{equation}
\sigma_{t+1}^2 = \omega + \alpha (r_t - \mu)^2 + \beta \sigma_t^2,
\label{eq:garch11}
\end{equation}
where $r_t$ denotes the log return at time $t$ and $\mu$ is the average return.
The regressive structure of Eq. \ref{eq:garch11}, whereby today's volatility depends on past volatility, captures the phenomenon of volatility clustering often observed in financial time series. Here, $\omega > 0$ is a constant term, $\alpha \geq 0$ captures the immediate impact of new shocks, and $\beta \geq 0$ reflects the persistence of volatility over time.
To ensure that volatility does not diverge, the model requires a stationarity condition:
\begin{equation}
\alpha + \beta < 1.
\end{equation}

GARCH is a widely used model for capturing time-varying volatility in financial time series.  By including GARCH(1,1) alongside SVR and QSVR, we establish our benchmarking against a standard econometric baseline. This provides a meaningful reference point for evaluating the potential gains of kernel-based and quantum-enhanced models in volatility forecasting.

\paragraph{SVR framework} In its linear form, SVR aims to find a function $f(x) = \mathbf{w}^\top \mathbf{x}_i+b$ that approximates the target $y$ while maintaining a margin of tolerance $\epsilon$ around the prediction. This margin is known as the $\epsilon$-insensitive tube: as long as the predicted value lies within $\pm \epsilon$ of the true $y_i$, no penalty is incurred.
Predictions that fall outside $\epsilon$ are those where the error exceeds the allowed tolerance, i.e., $|y_i - f(\mathbf{x}_i)| > \epsilon$. In such cases, the excess error is measured by the slack variables $\xi_i$ and $\xi_i^*$, which quantify how far above or below the tube the prediction lies. The optimization problem then balances regularization (small $|\mathbf{w}|^2$) against the total size of these violations:

\begin{equation}
\min_{\mathbf{w}, b} \ \frac{1}{2} |\mathbf{w}|^2 + C \sum_{i=1}^{N} \left( \xi_i + \xi_i^* \right),
\end{equation}

subject to:
\begin{equation}
\left\{
\begin{aligned}
& y_i - \mathbf{w}^\top \mathbf{x}_i - b \leq \epsilon + \xi_i, \\
& \mathbf{w}^\top \mathbf{x}_i + b - y_i \leq \epsilon + \xi_i^*, \\
& \xi_i, \, \xi_i^* \geq 0,
\end{aligned}
\right.
\end{equation}
where $C$ controls the trade-off between model complexity and margin violations, and $\xi_i, \xi_i^{*}$ are slack variables that allow soft violations of the margin. 
While linear SVR is effective when the relationship between input features and the target is approximately linear, it is limited in its ability to capture nonlinear dependencies \cite{smola2004}. To address this, kernel methods extend SVR by implicitly mapping the inputs into a higher-dimensional feature space. This is achieved through a kernel function $\kappa(x_i, x_j)$ which computes the inner product between feature representations in a high-dimensional feature space:
\begin{equation}
\kappa(x_i, x_j) = \langle \phi(x_i), \phi(x_j) \rangle.
\end{equation}

Referred to as \textbf{Kernel trick}, this formulation enables SVR to model nonlinear relationships in the original input space without ever explicitly computing the feature map $\phi(\cdot)$. In practice, one only needs to evaluate $\kappa(x_i, x_j)$, which makes the approach computationally efficient while greatly expanding the class of functions that can be learned.
Common kernel choices include the \textit{Radial Basis Function (RBF) kernel} and the \textit{polynomial kernel}:
\begin{equation}
\kappa_{\text{rbf}}(x_i, x_j) = \exp\left(-\gamma \|x_i - x_j\|^2\right),
\end{equation}

\begin{equation}
\kappa_{\text{poly}}(x_i, x_j) = \left(\gamma \, x_i^\top x_j + r \right)^d.
\end{equation}
In both RBF and polynomial kernels, $\gamma$ is the scaling factor. On the other hand, $r \ge 0$ is the offset term in the polynomial kernel, and $d$ is the degree of the polynomial kernel. These kernels allow the SVR algorithm to explore richer, higher-dimensional feature spaces in order to approximate more complex functions.

\begin{table*}[t]
\centering
\footnotesize
\renewcommand{\arraystretch}{1.25}
\caption{\textbf{Hyperparameter search space for classical and quantum SVR models}. For our QSVR, $\alpha$, $\beta$, and the entanglement topology are also hyperparameters. We retrain each model on the entire in‐sample window using the best hyperparameters and generate a one‐step‐ahead prediction $\hat{RV}_{t+1}$.}
\begin{tabular*}{\textwidth}{@{\extracolsep{\fill}} l c c c c c c @{}}
\toprule
\textbf{Model} & $C$ & $\epsilon$ & \textbf{degree} & $\gamma$ & \textbf{\# qubits} & \textbf{\# layers} \\
\midrule
$\text{SVR}_{\text{lin}}$  & $[10^{-2},\,10^2]$ & $[10^{-3},\,1]$ & --        & --                               & --              & --            \\
$\text{SVR}_{\text{poly}}$ & $[10^{-2},\,10^2]$ & $[10^{-3},\,1]$ & $\{2,3\}$ & $\{\text{scale},\,\text{auto}\}$ & --              & --            \\
$\text{SVR}_{\text{rbf}}$  & $[10^{-2},\,10^2]$ & $[10^{-3},\,1]$ & --        & $\{\text{scale},\,\text{auto}\}$ & --              & --            \\
QKSVR                      & $[10^{-2},\,10^2]$ & $[10^{-3},\,1]$ & --        & --                               & $\{6,8,10,12\}$ & $\{0,1,2,3\}$ \\
\bottomrule
\end{tabular*}
\label{table8}
\end{table*}

\begin{figure*}[b]
    \centering
    \includegraphics[width=0.9\textwidth]{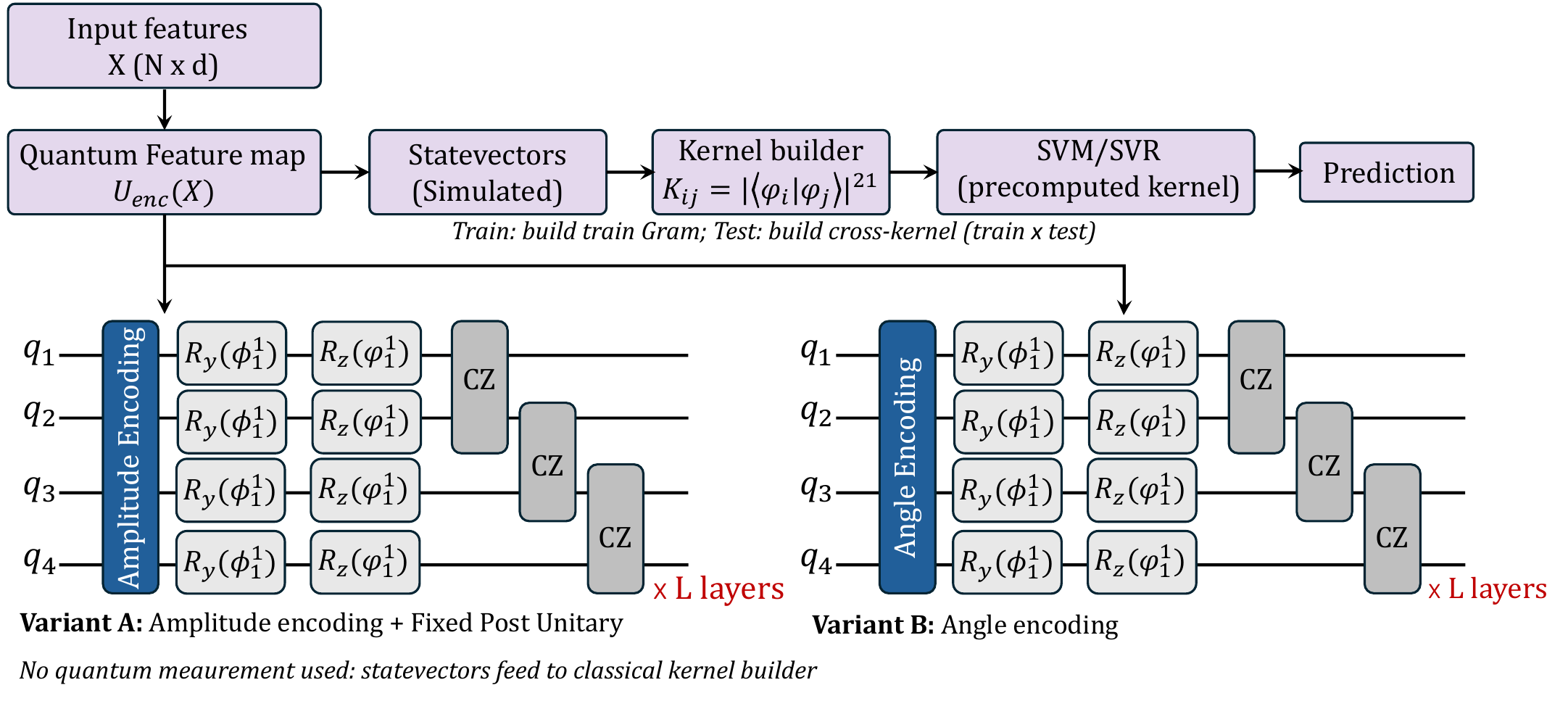}
    \caption{\scriptsize \textbf{QSVR pipeline and feature-map variants.} (a) Classical features are embedded by a quantum feature map. (b) The simulator evolves the circuit unitarily and outputs the full (c) \emph{simulated statevector} $\psi(x)$. (d) A kernel   $K_{ij}=|\langle\psi_i|\psi_j\rangle|^2$ is then computed from exact overlaps of these statevectors. (e) Subsequently, an SVR uses the precomputed kernel to make predictions. (b i) Variant A uses \emph{amplitude encoding} with a fixed, data-independent post-unitary repeated $L$ layers. (b ii) Variant B uses \emph{angle encoding} with CZ entanglers repeated $L$ layers.  In simulation, kernel entries are deterministic since they are built from exact statevector overlaps. On real quantum hardware, by contrast, such overlaps would need to be estimated through measurement-based fidelity tests (e.g., SWAP or Hadamard).}
    \label{fig:qsvr_pipeline}
\end{figure*}

\paragraph{QSVR framework}\mbox{}\\

 Quantum kernel methods exploit the exponentially large Hilbert space of $n$ qubits to capture high-order interactions among input features. Under this framework, a $d$-dimensional sample $x\in\mathbb R^d$ is embedded into a quantum state via the feature map $\Phi(x)\;=\;U(x)\,|0\rangle^{\otimes n}$. This constructs a nonlinear mapping onto $\mathbb C^{2^n}$ . Similar to classical kernel methods, the motivation for this mapping is that the inner product in this space: 
\begin{equation}
\kappa_q(x_i, x_j) = \left| \braket{\Phi(x_i) | \Phi(x_j)} \right|^2,
\end{equation}
 may capture the manifold of the data better than the original space. Following this, the SVR solves its dual quadratic program using the pre-computed kernel matrix, where the optimization depends only on inner products $\kappa(x_i,x_j)$ (classical or quantum).  Kernel evaluation scales by $\mathcal{O}(N^2)$ where $N$ is the number of samples.  Although rigorous proofs of quantum advantage remain an open problem, this enhanced capacity to encode complex covariance structures motivates our investigation of quantum kernels for volatility forecasting. Fig.~\ref{fig:qsvr_pipeline} summarizes the QSVR pipeline and the two feature-map variants used in our experiments.

The angle encoded map used in our study can be expressed as in Equation. \ref{eq:ang_enc_mp}, where each unitary block $U_{\text{rot}}$ consists of a hardware‐efficient $R_Y$–$R_Z$ rotations with a ring of CZ entanglers. This is repeated $L$ times (depth).

\begin{equation}\label{eq:ang_enc_mp}
\ket{\Phi_{\mathrm{rot}}(x)} = \bigl(U_{\mathrm{rot}}\, R_Y(x)^{\otimes d}\bigr)^L \ket{0}^{\otimes d},
\end{equation}

On the other hand, amplitude encoding embeds the entire normalized feature vector into the state amplitudes of $n=\lceil\log_2 d\rceil$ qubits. With a fixed, randomly sampled unitary layer stack $U_{rnd}^{L}$, the map is:
\begin{equation}
\Phi_{\rm amp}(x)  = U_{\text{rnd}}^{(L)}\bigl(\alpha \frac{x}{\|x\|}),   
\end{equation}
 where $\alpha$ is a normalization constant. To probe curvature in Hilbert space, we introduce an exponent $\beta \in \{1,2,3\}$ and evaluate: 
\begin{equation}
\kappa_q^{(\beta)}(x_i, x_j) = \left| \braket{\psi_i | \psi_j} \right|^{2\beta}.
\end{equation}

These design choices collectively provide a tunable framework for assessing the efficacy of quantum kernels in modeling financial-market volatility.

\subsubsection{Hyperparameter Search}
To find optimal configurations across models, we perform an extensive hyperparameter search using the Optuna framework\footnote{\url{https://github.com/optuna/optuna}} \cite{akiba2019optuna}. Optuna is a Bayesian optimization library that efficiently explores high-dimensional search spaces through techniques such as Tree-structured Parzen Estimators (TPE). For each SVR and QSVR variant, we define a model-specific search space, which is summarized in Table \ref{table8}.

\begin{table*}[b]
\centering
\footnotesize
\renewcommand{\arraystretch}{1.25}
\caption{Comparative performance of ANNs and QNNs on fold~5 using a low-dimensional (3-D) feature set. “Arch.” explicitly denotes the QNN variant (SQ, MQ, Hybrid-SQ, Hybrid-MQ). “Layers’’ indicates circuit depth, “Hyb.” denotes a classical pre-processing layer, and “MQR’’ refers to multi-qubit read-out.}
\begin{tabular*}{\textwidth}{@{\extracolsep{\fill}} l c c c c c
S[table-format=1.4]
S[table-format=1.4]
S[table-format=1.4]
S[table-format=1.4] @{}}
\toprule
\textbf{Ticker} & \textbf{Model} & \textbf{Arch.} & \textbf{Layers} & \textbf{Hyb.} & \textbf{MQR} &
\textbf{Accuracy} & \textbf{AUC} & \textbf{Precision} & \textbf{Recall} \\
\midrule
\texttt{GARAN.IS} & ANN & – & – & – & –     & 0.7625 & 0.8102 & 0.7921 & 0.6751 \\
                  & QNN & MQ  & 5 & False & True  & 0.6648 & 0.7093 & 0.6385 & 0.6721 \\[2pt]

\texttt{KCHOL.IS} & ANN & – & – & – & –     & 0.6747 & 0.7475 & 0.7294 & 0.6046 \\
                  & QNN & Hybrid-MQ & 6 & True  & True  & 0.7106 & 0.7968 & 0.7235 & 0.7262 \\[2pt]

\texttt{TCELL.IS} & ANN & – & – & – & –     & 0.7425 & 0.7597 & 0.7907 & 0.6693 \\
                  & QNN & SQ  & 4 & False & False & 0.6347 & 0.7430 & 0.7840 & 0.3858 \\[2pt]

\texttt{TUPRS.IS} & ANN & – & – & – & –     & 0.8323 & 0.9042 & 0.8233 & 0.8367 \\
                  & QNN & MQ  & 4 & False & True  & 0.7345 & 0.9039 & 0.6591 & 0.9469 \\[2pt]

\texttt{ULKER.IS} & ANN & – & – & – & –     & 0.7864 & 0.8760 & 0.7555 & 0.8381 \\
                  & QNN & MQ  & 3 & False & True  & 0.7525 & 0.8518 & 0.7860 & 0.6842 \\
\bottomrule
\end{tabular*}
\label{tab:qnn_lowdim_best}
\end{table*}

\subsubsection{Evaluation Metrics}
The standard metric of evaluation that we employ is Mean Squared Error (MSE), Mean Absolute Error (MAE), $R^2$, Directional Accuracy, and Quasi-likelihood (QLIKE). MSE measures the average of squared forecast errors, while MAE measures the average absolute difference between forecast and actual volatility. On the other hand, the coefficient of determination $R^2$ measures the share of variance in the realized volatility that our model’s forecasts explain. Values closer to 1 mean the model adds significant value beyond a naïve average. On the other hand, values near 0 mean we are no better than just using the long-run average

QLIKE is a volatility-specific metric and is defined as follows: 
\begin{equation}
\text{QLIKE}_{t} = \log(\hat{RV}_{t}) + \frac{RV_{t}}{\hat{RV}_{t}}.
\end{equation}

where $RV_t$ is the realized variance and $\hat{RV}_t$ its forecast.  Forecasts that are too low get punished more heavily than forecasts that are too high. This reflects that underestimating risk is often more dangerous. With QLIKE, \textit{lower scores indicate better forecasts}
Once we have the error series for our classical models and QSVR, we also carry out Diebold–Mariano Test to determine whether QSVRs provide better forecasts than their classical counterparts. 

\begin{table*}[t]
\centering
\footnotesize
\renewcommand{\arraystretch}{1.25}
\caption{Comparative performance of ANNs and QNNs on fold~5 for the \texttt{S\&P 500} dataset. 
The ANN baseline achieves slightly higher accuracy and AUC, while the Hybrid-SQ QNN (depth~3, Hybrid=True, MQR=False) shows competitive performance, trading off marginal accuracy for improved recall.}
\begin{tabular*}{\textwidth}{@{\extracolsep{\fill}} l c c c c c
S[table-format=1.4]
S[table-format=1.4]
S[table-format=1.4]
S[table-format=1.4] @{}}
\toprule
\textbf{Ticker} & \textbf{Model} & \textbf{Arch.} & \textbf{Layers} & \textbf{Hyb.} & \textbf{MQR} &
\textbf{Accuracy} & \textbf{AUC} & \textbf{Precision} & \textbf{Recall} \\
\midrule
\texttt{S\&P 500} & ANN & – & – & – & – &
\textbf{0.7289} & \textbf{0.7657} & 0.7529 & \textbf{0.7800} \\[2pt]

\texttt{S\&P 500} & QNN & Hybrid-SQ & 3 & True & False &
0.7221 & 0.7498 & \textbf{0.7319} & 0.8080 \\
\bottomrule
\end{tabular*}
\label{tab:qnn_middim_best}
\end{table*}

\section{Results and Discussion} \label{sec:ResultsDiscussion}

\subsection{Experimental Setup}
All experiments are implemented in Python. Quantum models are simulated using the \textit{TorchQuantum} library\footnote{\url{https://github.com/mit-han-lab/torchquantum}}~\cite{wang2022torchquantum}. Classical counterparts such as ANNs and LSTMs are built using PyTorch and Keras. For volatility forecasting, Support Vector Regression (SVR) models are implemented via \texttt{scikit-learn} while Quantum SVRs (QSVRs) are executed through custom TorchQuantum feature maps and a precomputed kernel matrix.
%
All quantum models are executed in simulation mode. Quantum circuits are evaluated using statevector simulation with exact expectation values. Random seeds are fixed at both NumPy and PyTorch levels to guarantee deterministic runs across folds.



\begin{table*}[t]
\centering
\footnotesize
\renewcommand{\arraystretch}{1.25}
\caption{Comparative performance of ANNs and QNNs on fold~5 for high-dimensional directional forecasting across five U.S. equities. QNNs achieve competitive results overall: for instance, the Hybrid-SQ QNN outperforms the ANN baseline on \texttt{AAPL} while the Hybrid-MQ QNN achieves a large recall gain on \texttt{DVN}. ANNs, however, retain the edge on accuracy and AUC for other tickers.}
\begin{tabular*}{\textwidth}{@{\extracolsep{\fill}} l c c c c c
S[table-format=1.3]
S[table-format=1.3]
S[table-format=1.3]
S[table-format=1.3] @{}}
\toprule
\textbf{Ticker} & \textbf{Model} & \textbf{Arch.} & $L$ & $n_{\mathrm{q}}$ & \textbf{MQR} &
\textbf{Accuracy} & \textbf{AUC} & \textbf{Precision} & \textbf{Recall} \\
\midrule
\texttt{AAPL} & ANN & – & – & – & –     
              & 0.572 & 0.597 & 0.587 & 0.613 \\
              & QNN & Hybrid-SQ & \textbf{6} & \textbf{4} & False &
                \textbf{0.606} & \textbf{0.635} & \textbf{0.614} & \textbf{0.663} \\[2pt]

\texttt{BA}   & ANN & – & – & – & –     
              & \textbf{0.677} & \textbf{0.738} & \textbf{0.680} & \textbf{0.701} \\
              & QNN & Hybrid-SQ & 3 & 5 & False &
                0.593 & 0.618 & 0.521 & 0.653 \\[2pt]

\texttt{GILD} & ANN & – & – & – & –     
              & 0.598 & 0.620 & 0.589 & 0.660 \\
              & QNN & Hybrid-SQ & \textbf{3} & \textbf{4} & False &
                \textbf{0.601} & 0.590 & 0.589 & \textbf{0.675} \\[2pt]

\texttt{DVN}  & ANN & – & – & – & –     
              & 0.651 & 0.675 & 0.662 & 0.665 \\
              & QNN & Hybrid-MQ & 5 & \textbf{4} & True &
                0.606 & 0.608 & 0.584 & \textbf{0.827} \\[2pt]

\texttt{LNC}  & ANN & – & – & – & –     
              & 0.627 & 0.694 & 0.653 & 0.647 \\
              & QNN & Hybrid-SQ & \textbf{6} & \textbf{5} & False &
                \textbf{0.625} & \textbf{0.663} & \textbf{0.641} & \textbf{0.681} \\
\bottomrule
\end{tabular*}
\label{tab:qnn_highdim_best}
\end{table*}

\subsection{Results for Directional Classification}

We now report the performance of QNNs and their classical counterpart on the same out-of-sample folds described in Fig.~\ref{fig:cv_splits}. All results refer to the final walk-forward test slice (fold 5) with early-stop and model selection performed as specified in Section~\ref{para:selection_criterion}.  
As introduced in Section~\ref{sec:Directional Classification} (see Fig.~\ref{fig:qnn_ann_arch}), four QNN architectures were explored: SQ, MQ, Hybrid-SQ, and Hybrid-MQ. Alongside the best-performing architecture, we also report circuit depth ($L$), qubit count ($n_q$), and whether multi-qubit readout (MQR) was applied.

\paragraph{Performance on 3-D Feature Set} Our analysis on the low-dimensional 3-D feature set reveals that across folds, QNNs demonstrate competitive performance to ANNs. Table~\ref{tab:qnn_lowdim_best} reports the detailed comparison between ANNs and QNNs on fold~5. While ANNs often achieve higher accuracy and AUC, QNNs are able to match or exceed them in certain cases, especially in the context of recall. This suggests that QNNs may capture signal structures differently, trading off small losses in overall accuracy for gains in sensitivity to positive cases. Notably, the best-performing QNNs on this feature set are architectures with deeper layers and Multi-Qubit Readout strategy employed.

\paragraph{Performance on 7-D Feature Set}
The performance comparison results for directional classification on 7-D feature set are presented in Table \ref{tab:qnn_middim_best}
We observe that, for the medium-dimensional experiment on S\&P 500 using seven cross-asset features, the \textit{classical ANNs model performs slightly better than QNNs} with the ANNs achieving higher AUC and accuracy. Similar to the case of low-dimensional (3-D) feature set, QNNs model posts a better recall. The Hybrid-SQ architecture with shallow depth and single-qubit readout is the best performing architecture on this feature set.

\paragraph{Performance on 64-D Feature Set}

The performance comparison results for directional classification on 64-D feature set are presented in Table \ref{tab:qnn_highdim_best}. In this regime,  QNNs remain competitive with classical ANNs. They attain a higher AUC than the ANN on AAPL with concomitant gains in accuracy. On the rest of the tickers, their performance is only marginally worse. Architecturally, deeper Hybrid-SQ circuits and Hybrid-MQ variants with MQR strategy tend to outperform pure QNN architectures.

\paragraph{Key Observations}

\begin{itemize}[leftmargin=1.5em]
    \item \textbf{Architectural Drivers:} Circuit depth, hybrid preprocessing, and readout design are the main determinants of QNN effectiveness. 
    \item \textbf{Depth–Hybrid Trade-off:} Both hybrid preprocessing and multi-qubit readout emerge as key design factors influencing QNN performance. Hybrid preprocessing enables shallower circuits to remain competitive with ANNs while multi-qubit readout enhances deeper architectures by improving their ability to capture complex feature interactions.
    \item \textbf{Hybrid QNN Gains:} Hybrid QNNs yield their strongest improvements in higher-dimensional feature spaces, outperforming parameter-matched ANNs on AAPL (+3.8 AUC / +3.4 pp accuracy) and KCHOL (+4.9 AUC / +3.6 pp accuracy), while achieving comparable accuracy on the remaining assets. 
    \item \textbf{Mid-dimensional Case:} In medium-dimensional tasks such as the 7-feature S\&P 500 experiment, ANNs maintain a slight edge in AUC and accuracy, while QNNs achieve higher recall.
\end{itemize}

\subsection{Results for Live Trading}

In order to evaluate whether predictive signals translate into real economic value, we assessed both LSTM and QLSTM models in a live trading simulation across four distinct market regimes of the S\&P 500. These regimes, which we label F1, F2, F3, and F4, are defined in Table \ref{tab:regimestats}. The detailed performance metrics—including AUC, annualized return, volatility, and risk-adjusted ratios are summarized in Table \ref{tab:trading_metrics}. Although the classifier skill is modest across both regimes for both of the models, the trading outcomes differed markedly once scores were thresholded into positions:

\begin{itemize}
    \item \textbf{F1 (Global Financial Crisis):} QLSTM improves risk-adjusted returns over LSTM despite similar performance on AUC (0.5446 vs.\ 0.5334). The Annualized return of QLSTM was $+5.96\%$ in comparison to LSTM's return of $-1.67\%$ to $+5.96\%$. This suggests the QLSTM scores align better with the sign and \emph{magnitude} of next-day moves in a high-volatility regime.

    \item \textbf{F2 (Pre-Covid):} QLSTM lifts ARC from $7.45\%$ to $11.03\%$ and reduces ASD by $\sim 40\%$ (0.150 $\rightarrow$ 0.088), yielding a Sharpe ratio of $1.26$ versus $0.50$.

    \item \textbf{F3 (Covid shock \& recovery):} QLSTM’s AUC is again higher ($+0.011$), but Sharpe is lower (0.766 vs.\ 0.895) as ASD rises (0.235 $\rightarrow$ 0.262). The Sortino ratio is slightly higher for QLSTM, implying a more favorable downside profile even as total volatility is larger. Returns of both QLSTM and LSTM are of similar magnitude, 20.06\% vs. 21.02\%, respectively.

    \item \textbf{F4 (Post-pandemic):} QLSTM underperforms economically (ARC $-15.15\%$, Sharpe $-0.694$) versus LSTM (ARC $35.16\%$, Sharpe $1.523$). ASD is marginally lower for QLSTM (0.231 $\rightarrow$ 0.218), so the shortfall is return-driven rather than risk-driven. This indicates that the fixed threshold calibration learned on the $\tau$-slice does not transfer well to this regime for QLSTM, even though its ranking (AUC) is slightly better.
\end{itemize}

\begin{table*}[t]
\centering
\footnotesize
\renewcommand{\arraystretch}{1.2}
\caption{Performance metrics for LSTM and QLSTM models across four folds. Metrics include Test AUC, Annualized Return (ARC), Annualized Standard Deviation (ASD), Sharpe ratio, Maximum Drawdown (MaxDD), and Sortino ratio.}
\label{tab:trading_metrics}
\begin{tabular*}{\textwidth}{@{\extracolsep{\fill}} c c c c c
S[table-format=1.4]
S[table-format=1.4]
S[table-format=1.4]
S[table-format=1.4]
S[table-format=1.4] @{}}
\toprule
\textbf{Fold} & \textbf{Model} & \textbf{hidden\_dim} & \textbf{depth} & \textbf{params} &
\textbf{Test AUC} & \textbf{ARC} & \textbf{ASD} & \textbf{Sharpe} & \textbf{Sortino} \\
\midrule
1 & LSTM  & 4 & 2 & 325 & 0.5334 & -0.0167 & 0.3492 & -0.0477 & -0.0661 \\
  & QLSTM & 4 & 2 & 233 & 0.5446 &  0.0596 & 0.1543 &  0.3865 &  0.2664 \\
2 & LSTM  & 4 & 2 & 325 & 0.5071 &  0.0745 & 0.1500 &  0.4966 &  0.6001 \\
  & QLSTM & 5 & 4 & 476 & 0.5170 &  0.1103 & 0.0876 &  1.2585 &  1.0413 \\
3 & LSTM  & 5 & 6 & 1426& 0.5184 &  0.2102 & 0.2349 &  0.8947 &  0.8125 \\
  & QLSTM & 5 & 3 & 386 & 0.5290 &  0.2006 & 0.2620 &  0.7659 &  0.8544 \\
4 & LSTM  & 4 & 4 & 645 & 0.5346 &  0.3516 & 0.2309 &  1.5228 &  3.0375 \\
  & QLSTM & 5 & 6 & 656 & 0.5429 & -0.1515 & 0.2184 & -0.6936 & -1.1116 \\
\bottomrule
\end{tabular*}
\end{table*}

\paragraph{Key Observations}
\begin{itemize}
    \item \textbf{Regime-specific Gains:} QLSTM improves risk-adjusted returns in two of four market regimes, namely the Global Financial Crisis (2008–09) and the Pre-COVID phase (2018–19), despite only modest AUC gains.
    \item \textbf{Calibration under Volatility:} The improved Sharpe ratios during these periods suggest QLSTM’s score calibration aligns closely with return magnitudes under volatile conditions. 
    \item \textbf{Threshold Sensitivity:} Performance is highly sensitive to threshold selection; Sharpe-maximizing long/short cutoffs significantly influence realized returns.
\end{itemize}

\begin{table*}[t]
\centering
\footnotesize
\renewcommand{\arraystretch}{1.25}
\caption{Comparative performance of QSVRs (angle encoding, 10 qubits) against classical SVRs and GARCH across four equities. Bold indicates the lowest QLIKE per ticker.}
\begin{tabular*}{\textwidth}{@{\extracolsep{\fill}} l l 
S[table-format=1.4]
S[table-format=1.2e-1]
S[table-format=1.4]
S[table-format=1.4]
l @{}}
\toprule
\textbf{Ticker} & \textbf{Model} & \textbf{QLIKE} & \textbf{MSE} & \textbf{$R^2$} & \textbf{DirAcc} & \textbf{DM–$p$} \\
\midrule
\multirow{5}{*}{\texttt{S\&P 500}}
  & SVR (Linear)  & -7.8748 & 1.71e-8 & -0.0164 & 0.5438 & $< 10^{-13}$
 \\
  & SVR (Poly)    & -8.0569 & 1.40e-8 & 0.1671  & 0.5424 & $<10^{-13}$ \\
  & SVR (RBF)     & \textbf{-8.3277} & 0.88e-8 & 0.4734  & 0.6245 & 0.0015 \\
  & GARCH         & -8.2488 & 0.82e-8 & 0.5106  & 0.5382 & 0.0335  \\
  & QSVR (10q)    & -8.2937 & 1.43e-8 & 0.1482  & 0.6203 & — \\
\midrule
\multirow{5}{*}{\texttt{AAPL}}
  & SVR (Linear)  & -7.2522 & 6.65e-8 & 0.3072  & 0.5508 & $<10^{-9}$ \\
  & SVR (Poly)    & -7.3288 & 3.17e-8 & 0.6697  & 0.5605 & 0.0042 \\
  & SVR (RBF)     & \textbf{-7.4101} & 2.73e-8 & 0.7152  & 0.6467 & 0.0010  \\
  & GARCH         & -7.2831 & 5.04e-8 & 0.4752  & 0.5327 & $2\times10^{-5}$  \\
  & QSVR (10q)    & -7.3888 & 3.27e-8 & 0.6595  & 0.6287 & —  \\
\midrule
\multirow{5}{*}{\texttt{GARAN}}
  & SVR (Linear)  & -6.0486 & 6.61e-4 & -447.62 & 0.5355 & $<10^{-15}$  \\
  & SVR (Poly)    & 60.5361 & 3.13e-6 & -1.1245 & 0.5605 & 0.277  \\
  & SVR (RBF)     & \textbf{-6.2196} & 7.56e-7 & 0.4865  & 0.6273 & 0.145  \\
  & GARCH         & -6.1398 & 8.68e-7 & 0.4105  & 0.5800 & 0.006  \\
  & QSVR (10q)    & -6.2121 & 7.89e-7 & 0.4643  & 0.6008 & —  \\
\midrule
\multirow{5}{*}{\texttt{KCHOL}}
  & SVR (Linear)  & -6.2930 & 9.58e-4 & -1096.91 & 0.5382 & $<10^{-15}$  \\
  & SVR (Poly)    & -6.3912 & 1.60e-5 & -17.3024 & 0.5828 & 0.009  \\
  & SVR (RBF)     & -6.4247 & 4.87e-7 & 0.4421   & 0.6481 & 0.095  \\
  & GARCH         & -6.3593 & 5.94e-7 & 0.3193   & 0.5897 & 0.0003  \\
  & QSVR (10q)    & \textbf{-6.4352} & 3.88e-7 & 0.5558   & 0.6328 & —  \\
\bottomrule
\end{tabular*}
\label{tab:qsvr_angle_all}
\end{table*}

\begin{table*}[t]
\centering
\footnotesize
\renewcommand{\arraystretch}{1.25}
\caption{Comparative performance of QSVRs (amplitude encoding, $d=32$, 5 qubits) against classical SVRs and GARCH across four equities. Bold indicates the lowest QLIKE per ticker.}
\label{tab:amp_qsvm_all}
\begin{tabular*}{\textwidth}{@{\extracolsep{\fill}} l l 
S[table-format=1.4]
S[table-format=1.2e-1]
S[table-format=1.4]
S[table-format=1.4]
l @{}}
\toprule
\textbf{Ticker} & \textbf{Model} & \textbf{QLIKE} & \textbf{MSE} & \textbf{$R^2$} & \textbf{DirAcc} & \textbf{DM–$p$} \\
\midrule
\multirow{5}{*}{\texttt{SPX}}
  & SVR (Linear)  & -7.9012 & 1.53e-8 & 0.0871   & 0.5563 & $1.3\times10^{-7}$ \\
  & SVR (Poly)    & -8.0246 & 1.89e-8 & -0.1282  & 0.5577 & $1.0\times10^{-6}$ \\
  & SVR (RBF)     & \textbf{-8.2845} & 8.05e-8 & -3.7953  & 0.5855 & 0.00037 \\
  & GARCH         & -8.2488 & 8.22e-9 & 0.5106   & 0.5382 & 0.0199 \\
  & QSVM (5q)     & -8.1824 & 1.27e-8 & 0.2424   & 0.5786 & — \\
\midrule
\multirow{5}{*}{\texttt{AAPL}}
  & SVR (Linear)  & -7.2960 & 1.81e-7 & -0.8834  & 0.5814 & 0.656  \\
  & SVR (Poly)    & -7.2926 & 8.58e-8 & 0.1062   & 0.5675 & 0.540  \\
  & SVR (RBF)     & \textbf{-7.3876} & 1.44e-7 & -0.5006  & 0.6008 & 0.00015 \\
  & GARCH         & -7.2824 & 5.04e-8 & 0.4750   & 0.5299 & 0.333  \\
  & QSVM (5q)     & -7.3065 & 5.38e-8 & 0.4394   & 0.5911 & — \\
\midrule
\multirow{5}{*}{\texttt{GARAN}}
  & SVR (Linear)  & -6.0734 & 2.57e-3 & -1743.97 & 0.5605 & 0.0233 \\
  & SVR (Poly)    & -5.9171 & 3.25e-6 & -1.2056  & 0.5577 & 0.119  \\
  & SVR (RBF)     & -6.1386 & 1.11e-6 & 0.2482   & 0.5953 & 0.0062 \\
  & GARCH         & \textbf{-6.1391} & 8.70e-7 & 0.4095   & 0.5828 & 0.0119 \\
  & QSVM (5q)     & -5.6154 & 1.30e-6 & 0.1199   & 0.5800 & — \\
\midrule
\multirow{5}{*}{\texttt{KCHOL}}
  & SVR (Linear)  & -6.2909 & 9.75e-5 & -110.72  & 0.5480 & 0.0015 \\
  & SVR (Poly)    & -6.2705 & 5.81e-6 & -5.6608  & 0.5925 & 0.0092 \\
  & SVR (RBF)     & \textbf{-6.3723} & 6.30e-7 & 0.2782   & 0.6398 & $2.4\times10^{-9}$ \\
  & GARCH         & -6.3586 & 5.93e-7 & 0.3208   & 0.5925 & $1.2\times10^{-5}$ \\
  & QSVM (5q)     & -6.1804 & 6.61e-7 & 0.2423   & 0.5828 & — \\
\bottomrule
\end{tabular*}

\end{table*}

\subsection{Results for Volatility Forecasting}

We now present the results for volatility forecasting, summarized in Tables~\ref{tab:qsvr_angle_all} and \ref{tab:amp_qsvm_all}. Unlike the directional classification task, volatility estimation requires capturing second-order dynamics and long memory in return series, which makes kernel-based methods a natural benchmark. Similar to our Directional Classification study \ref{sec:Directional Classification}, we investigate two distinct encoding techniques for the QSVR. While angle encoding allows us to embed low- and medium-dimensional inputs into quantum circuits, amplitude encoding, by contrast, enables the inclusion of a larger set of lagged variables, which allows us to test whether incorporating longer historical windows improves volatility forecasts.

\paragraph{Angle Encoding Results}  
Table~\ref{tab:qsvr_angle_all} shows that QSVRs with angle embedding consistently outperform linear and polynomial SVRs on QLIKE (positive DM statistics) across all four tickers, though they still lag the RBF kernel. This indicates that quantum kernels are indeed expressive enough to capture nonlinear variance clustering, but their inductive bias is not yet as strong as the classical RBF. Importantly, DirAcc remains stable around $0.60$, often on par with or slightly below the best classical models. This suggests that even when QLIKE values are worse than RBF, QSVR forecasts correctly identify the \emph{direction} of variance changes. In other words, they capture clustering effects in volatility paths even if absolute levels are imperfectly calibrated. On S\&P 500 and KCHOL, for instance, QSVR achieves the strongest balance between QLIKE and DirAcc, underscoring their viability in medium-sample regimes.

\paragraph{Amplitude Encoding Results}  
Table~\ref{tab:amp_qsvm_all} illustrates the more delicate behavior of amplitude-encoded QSVRs. With $d=32$, the model attains competitive QLIKE performance, but increasing dimensionality to $d=64$ results in clear degradation: the forecasts lose accuracy in estimating the mean and variance of returns. This sensitivity is expected because amplitude encoding requires precise state preparation, and numerical instability compounds as the embedding dimension grows. Nevertheless, DirAcc remains at competitive levels ($\sim0.58$–$0.60$), meaning the model still tracks the \emph{sign} of volatility shifts. These findings imply that while amplitude encoding has theoretical advantages in compressing high-dimensional features, practical implementations must carefully balance $d$ against circuit depth and shot noise. 


\subsubsection{ Key Observation}  
Integrating the insights from both angle and amplitude encoding experiments, the volatility forecasting results reveal the following consistent patterns:
\begin{itemize}[leftmargin=1.5em]
    \item \textbf{Relative Positioning:} QLIKE values consistently position QSVRs between RBF SVRs (best) and linear/polynomial SVRs (worst). This indicates they are a viable intermediate kernel option. 
    \item \textbf{Explanatory Power:} $R^2$ values show that QSVRs offer moderate explanatory power in comparison to classical RBF SVRs which still dominate in absolute fit. 
    \item \textbf{Directional Advantage:} QSVRs frequently match or slightly exceed classical models in directional accuracy (DirAcc).
    \item \textbf{Empirical Strengths:} QSVR achieves the lowest QLIKE on KCHOL and performs closely to RBF SVR on AAPL and S\&P 500, demonstrating competitiveness in moderate-sample, higher-dimensional settings where quantum embeddings capture nonlinear dependencies.
\end{itemize}


In the amplitude encoding regime ($d=16, 64, 32$), QSVR delivers best performance when $d=32$. When $d$ is pushed to 64, their performance degrades, and their ability to estimate the mean of the data suffers.

\section{Conclusion} \label{sec:Conclusion}

This study presented a unified benchmark to compare quantum and classical models across three core equity forecasting tasks: directional classification, live trading, and volatility forecasting. By aligning data splits, architectures, and parameter budgets, we ensured fair comparisons between models and reported
performance on consistent evaluation metrics.
Our results show that quantum models can compete with or surpass classical approaches in certain settings. Shallow QNNs perform well on compact features, amplitude encoding enables useful signal extraction from higher-dimensional data, QLSTMs improve
risk-adjusted returns in multiple regimes, and QSVRs reliably match or exceed linear SVRs. At the same time, classical models retain an advantage on mid-range cross-asset features, showing that the strengths of classical and quantum models are context-dependent.
Beyond the empirical findings, our work contributes a standardized benchmark for evaluating QML in finance. It highlights trade-offs between encoding schemes, circuit depth, readout choices, and hybrid designs, linking them to practical outcomes such as accuracy, drawdown, and Sharpe ratios. These insights offer a clearer picture of when and why quantum approaches add value. Our framework can serve as a foundation for future studies exploring more advanced quantum architectures, noise-resilient training strategies, and hybrid classical–quantum designs. As quantum hardware matures, we expect the intersection of quantum computation and financial modeling to yield novel insights into both algorithmic design and practical trading applications.

 \section*{Acknowledgment}
 This work was supported in part by the NYUAD Center for Quantum and Topological Systems (CQTS), funded by Tamkeen under the NYUAD Research Institute grant CG008.

\bibliographystyle{unsrt}
\bibliography{references}

\newpage
\begin{IEEEbiography}[{\includegraphics[width=1.0in,height=1.10in]{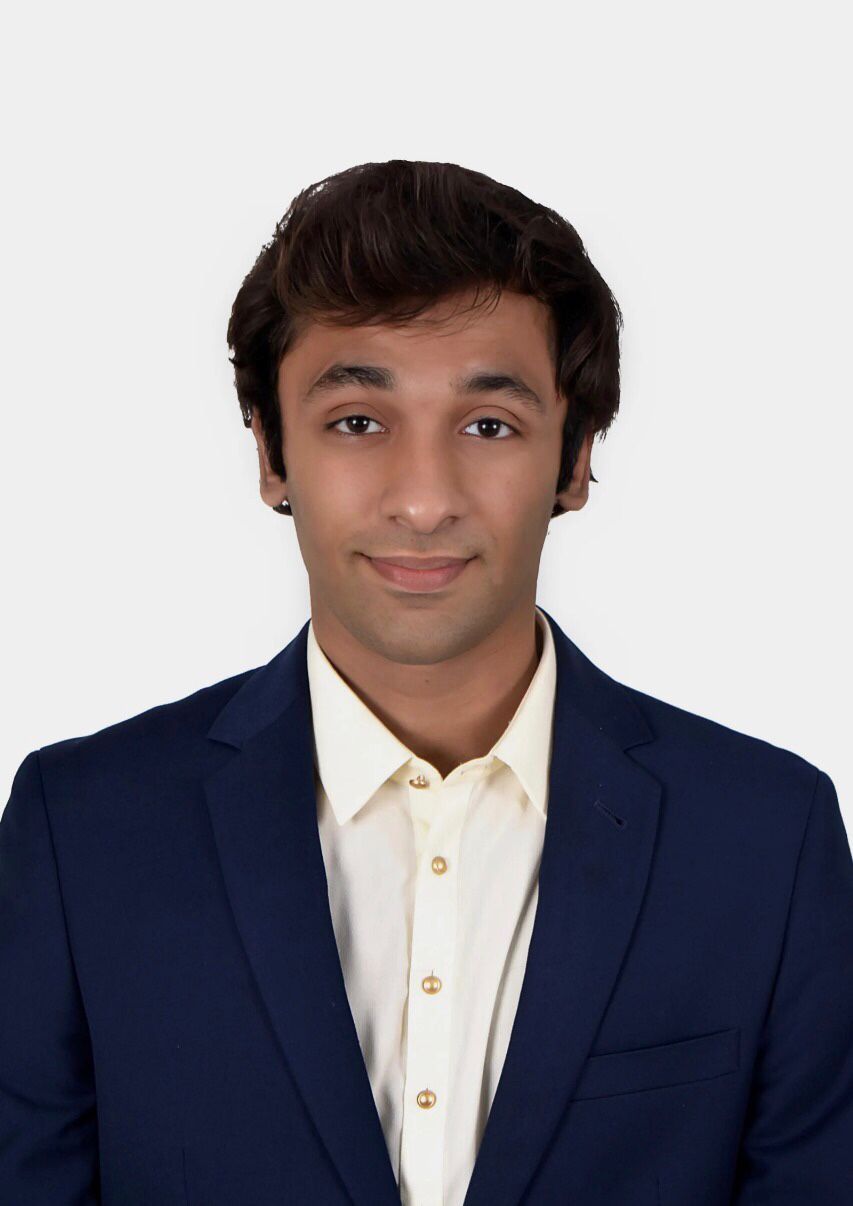}}]{Rehan Ahmad} received his Bachelors degree in Physics from the Lahore University of Management Sciences (LUMS), Lahore, Pakistan, in 2024, with a focus on quantum mechanics. His research interests include quantum machine learning for financial time-series forecasting, variational quantum circuits, and hybrid quantum–classical modeling.
\end{IEEEbiography}

\begin{IEEEbiography}[{\includegraphics[width=1.10in,height=1.25in]{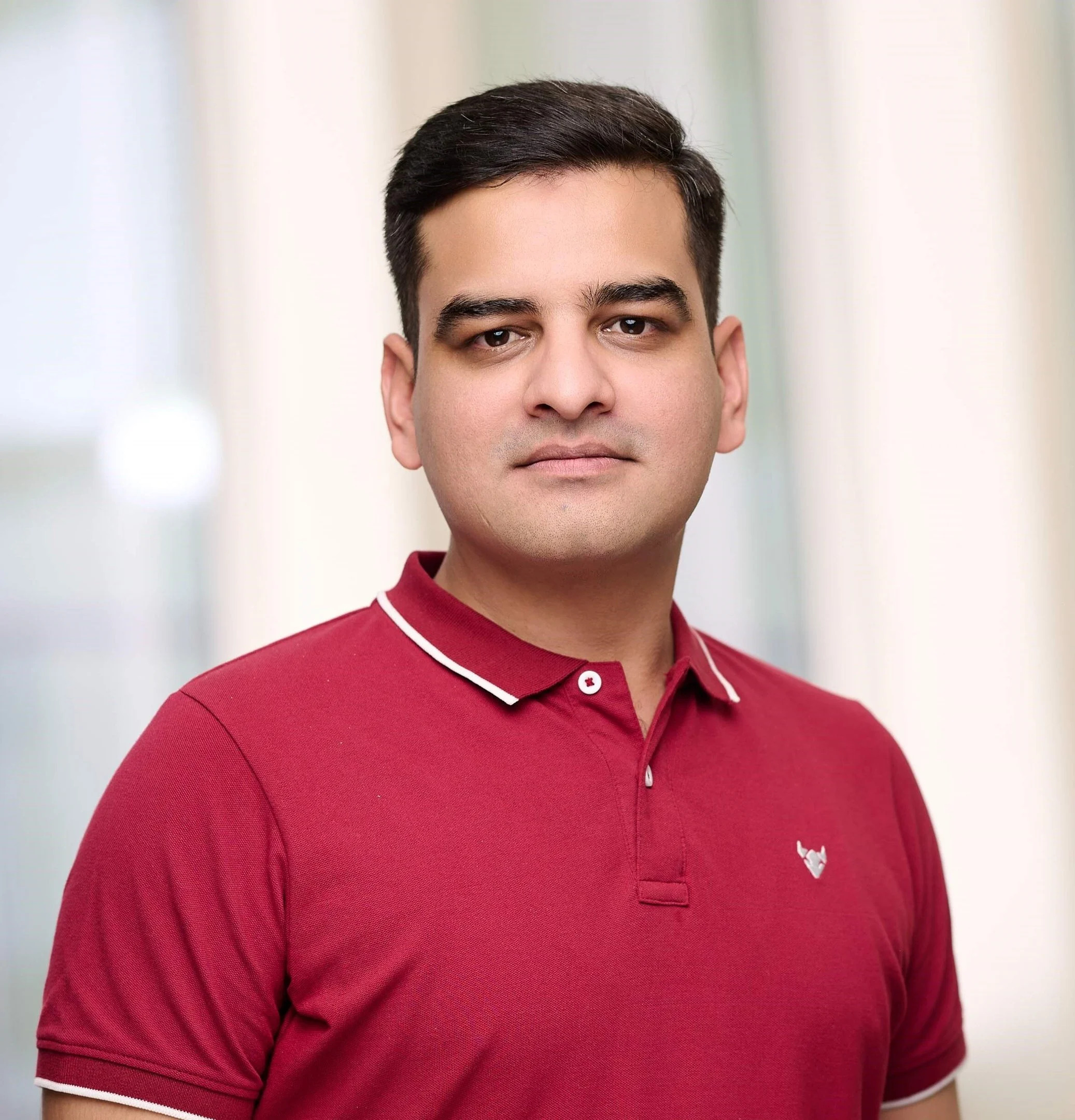}}]{Muhammad Kashif} obtained his Bachelor's degree in Electrical (Electronics) Engineering from COMSATS Institute of Information Technology, Pakistan, in 2015, followed by a Master’s degree in Electronics and Computer Engineering from Istanbul Sehir University, Turkey, in 2020. He then completed his Ph.D. at Hamad Bin Khalifa University, Qatar, focusing on exploring potential quantum advantages in machine learning during the Noisy Intermediate-Scale Quantum (NISQ) era.
At present, Kashif serves as a Postdoctoral Researcher at the Center for Quantum and Topological Systems (CQTS) at New York University Abu Dhabi. His research interests encompass Quantum Machine Learning (QML) and classical machine learning, with a particular focus on the intersection of these fields and their mutual enhancement. He is also interested in the behavioral analysis of various QML algorithms on the near-term nosiy quantum devices and the development of noise-resilient techniques to enhance the trainability of QML algorithms. 
\end{IEEEbiography}

\begin{IEEEbiography}
[{\includegraphics[width=1.05in,height=1.95in,clip,keepaspectratio]{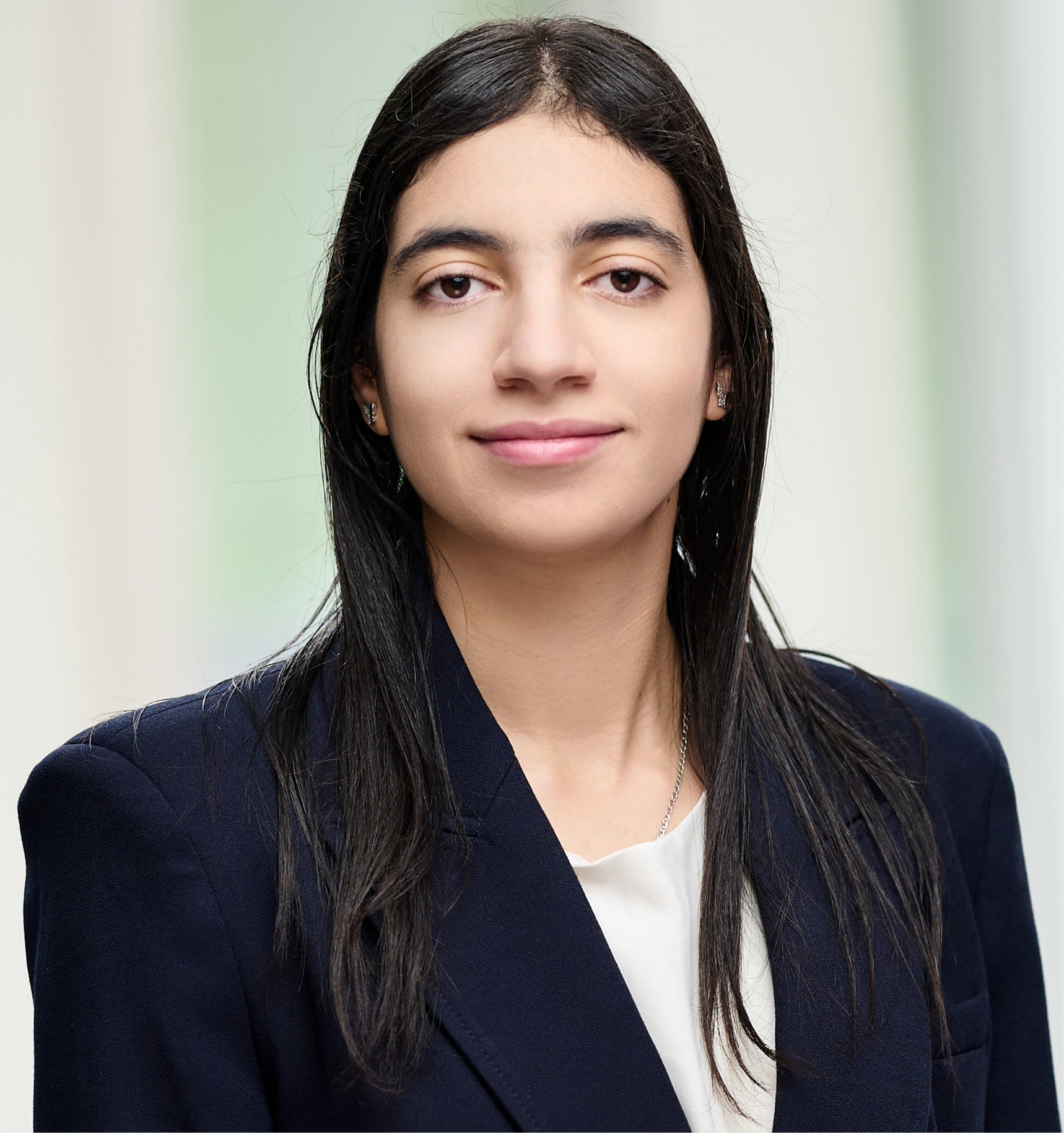}}]{Nouhaila Innan} is a Postdoctoral Associate at the Center for Quantum and Topological Systems (CQTS) and a Research Team Lead at the eBRAIN Lab, New York University Abu Dhabi. She earned her B.Sc. in Physics and Applications and M.Sc. in Physics and New Technologies, specializing in materials and nanomaterials, at Hassan II University of Casablanca, where she completed a PhD in Quantum Machine Learning in July 2024. Her research integrates principled advances in QML theory with systems-level engineering, encompassing algorithmic design, hardware-aware optimization, and end-to-end deployment. It systematically addresses fundamental limitations, device noise, data heterogeneity, privacy, and reproducible evaluation through rigorous benchmarking and application-driven validation. Beyond research, she mentors students and supports international outreach, helping translate complex quantum concepts into accessible, practice-oriented guidance.

\end{IEEEbiography}
\begin{IEEEbiography}[{\includegraphics[width=1in,height=1.25in,clip,keepaspectratio]{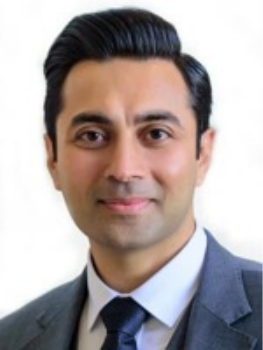}}]{Muhammad Shafique} (Senior Member,
IEEE) received the Ph.D. degree in computer
science from Karlsruhe Institute of Technology
(KIT), Germany, in 2011. Afterwards, he established and led a highly recognized research group with KIT for several years and conducted
impactful collaborative research and development activities across the globe. In October 2016, he joined the Institute of Computer Engineering, Faculty of Informatics, Technische Universität Wien (TU Wien), Vienna, Austria, as a Full Professor of computer architecture and robust, energy-efficient technologies. Since September 2020, he has been with New York University (NYU) Abu Dhabi, United Arab Emirates, where he is currently a Full Professor and the Director of
the eBrain Laboratory, and a Global Network Professor with the Tandon
School of Engineering, NYU-New York City, USA. He is also a CoPI/an
Investigator in multiple NYUAD Centers, including the Center of Artificial Intelligence and Robotics (CAIR), the Center of Cyber Security (CCS), the Center for Interacting Urban Networks (CITIES), and the Center for Quantum and Topological Systems (CQTS). He holds one U.S. patent, and has co-authored six books, more than ten book chapters, more than 350 papers in premier journals and conferences, and more than 100 archive articles. His research interests include AI and machine learning hardware and system-level design, brain-inspired computing, machine learning security and privacy, quantum machine learning, cognitive autonomous systems, wearable healthcare, energy-efficient systems, robust computing, hardware security, emerging technologies, FPGAs, MPSoCs, and embedded systems. He has given several keynotes, invited talks, and tutorials, and organized many special sessions at premier venues. He is a Senior Member of the IEEE Signal Processing Society (SPS) and a member of the ACM, SIGARCH, SIGDA, SIGBED, and HIPEAC. He has served as the PC chair, the general chair, the track chair, and a PC member for several prestigious IEEE/ACM conferences. He received the 2015 ACM/SIGDA Outstanding New Faculty Award, the AI 2000 Chip Technology Most Influential Scholar Award, in 2020 and 2022, the AI 2000 Chip Technology Most Influential Scholar Award, in 2020, 2022, and 2023, the ASPIRE AARE Research Excellence Award, in 2021, six gold medals, and several best paper awards and nominations at prestigious conferences.

\end{IEEEbiography}

\end{document}